\documentclass[sigconf]{acmart}
\AtBeginDocument{%
  }

\usepackage{command}
\usepackage{algorithm}
\usepackage{algpseudocode}

\setcopyright{acmlicensed}
\copyrightyear{2025}
\acmYear{2025}
\acmDOI{XXXXXXX.XXXXXXX}
\acmConference[MM '25]{33rd ACM International Conference on Multimedia}{October 27--31, 2025}{Dublin, Ireland}
\acmISBN{978-1-4503-XXXX-X/2018/06}

\begin{document}

\title{Disrupting Semantic and Abstract Features for Better Adversarial Transferability}

\author{Yuyang Luo}
\affiliation{%
  \institution{Brown University}
  \city{Providence}
  \country{USA}}

\author{Xiaosen Wang}
\affiliation{%
  \institution{Huazhong University of Science and Technology}
  \city{Wuhan}
  \country{China}}

\author{Zhijin Ge}
\affiliation{%
   \institution{Xidian University}
   \city{Xi'an}
  \country{China}}

\author{Yingzhe He}
\affiliation{%
   \institution{University of the Chinese Academy of Sciences}
   \city{Beijing}
  \country{China}}
  

\renewcommand{\shortauthors}{Luo et al.}

\begin{abstract}
  Adversarial examples pose significant threats to deep neural networks (DNNs), and their property of transferability in the black-box setting has led to the emergence of transfer-based attacks, making it feasible to target real-world applications employing DNNs. Among them, feature-level attacks, where intermediate features are perturbed based on feature importance weight matrix computed from transformed images, have gained popularity. In this work, we find that existing feature-level attacks primarily manipulate the semantic information to derive the weight matrix. Inspired by several works that find CNNs tend to focus more on high-frequency components (\aka abstract features, \eg, texture, edge, \etc), we validate that transforming images in the high-frequency space also improves transferability. Based on this finding, we propose a balanced approach called \fname. Specifically, \name conducts \textsc{BlockMix} on the input image and \textsc{Self-Mix} on the frequency spectrum when computing the weight matrix to highlight crucial features. By using such a weight matrix, we can direct the attacker to disrupt both semantic and abstract features, leading to improved transferability. Extensive experiments on the ImageNet dataset also demonstrate the effectiveness of our method in boosting adversarial transferability.
\end{abstract}






\maketitle

\section{Introduction}
\label{sec:intro}
Deep neural networks (DNNs) have achieved great performance in various computer vision tasks, such as image classification~\citep{simonyan2015very,he2016deep,szegedy2016rethinking}, objective detection~\citep{girshick2015fast, ren2015faster, he2017mask}, and autonomous driving~\citep{yu2022dual, wu2023transformation, wu2023policy}. However, recent works have shown that DNNs are vulnerable to adversarial examples~\citep{szegedy2014intriguing, goodfellow2015explaining}, in which applying human-imperceptible perturbations on clean input can cause misclassification. Furthermore, adversarial examples exhibit a compelling property known as transferability~\citep{dong2018boosting, lin2020nesterov}, whereby adversarial inputs crafted for a surrogate model are also capable of misleading other, independently trained victim models. This phenomenon significantly heightens the practicality of adversarial attacks in real-world settings, as it enables adversaries to compromise target models without requiring explicit knowledge of their internal architecture or parameters, thereby posing substantial security threats.


\begin{figure}
    \centering
    \begin{subfigure}{0.18\linewidth}
        \includegraphics[width=\linewidth]{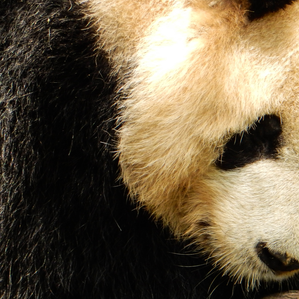}
        \caption*{Raw}
    \end{subfigure}
    \begin{subfigure}{0.18\linewidth}
        \includegraphics[width=\linewidth]{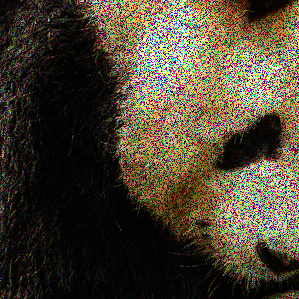}
        \caption*{FIA}
    \end{subfigure}
    \begin{subfigure}{0.18\linewidth}
        \includegraphics[width=\linewidth]{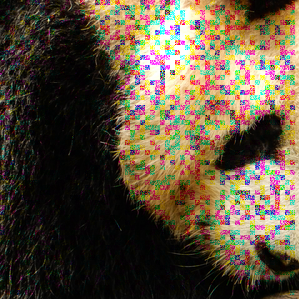}
        \caption*{RPA}
    \end{subfigure}
    \begin{subfigure}{0.18\linewidth}
        \includegraphics[width=\linewidth]{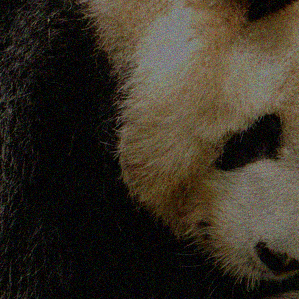}
        \caption*{NAA}
    \end{subfigure}
    \begin{subfigure}{0.18\linewidth}
        \includegraphics[width=\linewidth]{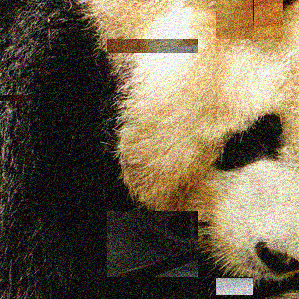}
        \caption*{\name}
    \end{subfigure}
    \vspace{-1em}
    \caption{The raw image and the transformed images for weight matrix calculation (not adversarial examples) in FIA, RPA, NAA, and our \name.}
    \label{fig:transformed_images}
    \vspace{-1em}
\end{figure}

Though several existing attack methods~\citep{kurakin2017adversarial,madry2018towards} have shown impressive attack performance in the white-box setting, their effectiveness notably diminishes in the black-box scenarios, particularly against models equipped with advanced defenses~\citep{kurakin2017adversarialb, kurakin2018ensemble}. Recently, numerous methods have been proposed to improve adversarial transferability, such as introducing momentum into gradient iterations~\citep{dong2018boosting, lin2020nesterov, wang2021enhancing, wang2021boosting, ge2023boosting}, adopting various input transformations~\citep{xie2019improving, wang2021admix, wang2023structure, wang2023boosting}, and integrating multiple models for attacks~\citep{liu2017delving,chen2023rethinking}. However, there is still a distinct gap between the performance of white-box and transfer-based black-box attacks.

Besides, there has been an increasing interest in feature-level attacks~\citep{wang2021feature, zhang2022enhancing, zhang2022improving}. These methods perturb the intermediate-level features based on the feature importance weight matrix to disrupt crucial object-aware features that predominantly influence model decisions for more transferable adversarial examples. The feature importance weight matrix is an aggregated gradient, which is obtained by calculating the average gradient \wrt the intermediate-level features of transformed images. We notice that transformations in those methods primarily modify the semantic information in the image space by masking or scaling to obtain the weight matrix. From the frequency-domain perspective, most of the critical semantic information is located in the low-frequency domain~\citep{wallace1991jpeg, guo2018low}. On the other hand, several studies~\citep{wang2020high, yin2019fourier, wang2020frequencybased} find that CNNs tend to focus more on high-frequency components, which pertain to abstract information such as image textures and edges. Such inconsistency among the attackers and models might limit the adversarial transferability, which inspires us to explore jointly disrupting the semantic and abstract features to further boost adversarial transferability.

Based on the above analysis, we hypothesize and empirically substantiate that manipulating high-frequency information to generate the weight matrix is also beneficial for improving transferability. Based on this finding, we propose a balanced attack called \fname. Specifically, \name transforms the images into the frequency domain to conduct \textsc{Self-Mix} process, where the frequency spectrum is mixed in a certain ratio with its randomly center-rotated one.
Since the low-frequency components occupy less portion in the frequency domain, this may result in suboptimal perturbation of semantic features. To mitigate this, \name randomly mixes the image blocks from other categories to the original image prior to the \textsc{Self-Mix} process, thereby further distorting the original semantic content. These dual transformations aim to strike a balance in emphasizing both semantic and abstract features when identifying crucial features. Subsequently, the weight matrix is calculated using the transformed images to establish a new objective function to highlight the crucial features for attacks to achieve better transferability. As in Fig.~\ref{fig:transformed_images}, compared with existing feature-level attacks, \name breaks the semantic features locally and abstract features uniformly yet unnoticeably. This helps identify more significant features for boosting adversarial transferability. Our contributions are summarized as follows:
\begin{itemize}
    \item We postulate and empirically validate that disrupting high-frequency components is beneficial for enhancing adversarial transferability.
    \item Based on this finding, we propose \name to disrupt both semantic and abstract features, overcoming the drawbacks of existing methods and enhancing the transferability of crafted adversarial examples.
    \item Extensive experiments on the ImageNet dataset demonstrate that \name can enhance transferability against both normally trained, adversarially trained, and defense models. Our approach outperforms the baseline methods, highlighting the effectiveness and superiority of \name
\end{itemize}

\section{Related Works}
\subsection{Adversarial Attacks.}
Generally, adversarial attacks can be divided into two categories, \ie, white-box attacks and black-box attacks. In the white-box setting, the attacker has all the information about the architecture and parameters of the target model~\citep{goodfellow2015explaining, kurakin2017adversarial}. By contrast, black-box attacks are more practical since they only access limited or no information about the target model. There are two types of black-box attacks~\citep{gao2021staircase,wang2022boosting,chen2020HopSkipJumpAttack}: query-based and transfer-based attacks. Query-based attacks~\citep{ilyas2018black,shi2019curls} often take hundreds or even thousands of queries to craft adversaries, making them inefficient. On the other hand, transfer-based attacks~\citep{dong2019evading, lin2020nesterov} generate adversaries on the surrogate model without accessing the target model, leading to superior practical applicability and attracting increasing attention.

While methods like I-FGSM~\citep{kurakin2017adversarial} are effective in white-box settings, they struggle with low transferability in black-box scenarios. To improve transferability, various approaches have been developed. Gradient-based attacks improve optimization techniques for better transferability. For instance, MIM~\citep{dong2018boosting} adds momentum into I-FGSM to stabilize updates and avoid poor local maxima. EMI~\citep{wang2021boosting} improves momentum by accumulating gradients from multiple data points aligned with the previous gradient. Inspired by data augmentation strategies~\citep{zhang2018mixup,verma2019manifold,yun2019cutmix}, input transformation-based attacks have been introduced. Xie~\etal~\cite{xie2019improving} use random resizing and padding to generate more transferable adversarial examples. Wang~\etal~\cite{wang2021admix} mix images from different categories with the original input while keeping the original label. Long~\etal~\cite{long2022frequency} transforms input images in the frequency domain. Besides, architecture-related approaches modify the source model to enhance transferability. SGM~\citep{Wu2020Skip} increases gradient backpropagation from skip connections in ResNet by adjusting the decay factor. LinBP~\citep{guo2020backpropagating} modifies ReLU gradients and rescales gradients in each block. BPA~\citep{wang2023rethinking} recovers the gradient truncated by non-linear layers using non-zero function.

Several approaches have been developed to enhance adversarial transferability by perturbing intermediate features. TAP~\citep{zhou2018transferable} increases the distance among intermediate features and smooths perturbations with a regularizer. ILA~\citep{huang2019enhancing} fine-tunes adversarial examples generated by methods like MIM by enhancing the feature difference between the original and adversarial examples at specific layers. FIA~\citep{wang2021feature} disrupts object-aware features by minimizing a weighted feature map, where the weights are derived from gradients of randomly pixel-wise masked input images. RPA~\citep{zhang2022enhancing} similarly uses averaged gradients from randomly masked images of varying patch sizes to emphasize key object-related features. NAA~\citep{zhang2022improving} estimates neuron importance using neuron attribution, combining positive and negative attributions to generate adversarial examples more efficiently. These methods typically disrupt semantic information but may overlook abstract features, potentially limiting their ability to boost transferability.

\subsection{Adversarial Defenses}
To mitigate adversarial threats, numerous defense strategies have been proposed. One of the most effective is adversarial training~\cite{goodfellow2015explaining, madry2018towards}, which improves model robustness by incorporating adversarial examples during training. Ensemble adversarial training~\cite{Tram2018ensemble} further enhances robustness by leveraging adversarial examples generated from other pre-trained models, demonstrating strong performance against transfer-based attacks. However, adversarial training often suffers from high computational cost and poor scalability, especially with large datasets and complex architectures. Therefore, several alternative defenses are proposed aiming to purify adversarial inputs before inference. JPEG compression~\cite{karolina2016study} can partially remove perturbations, while HGD~\cite{liao2018defense} learns a denoiser guided by high-level representations. NIPS-r3\footnote{https://github.com/anlthms/nips-2017/tree/master/mmd} applies input transformations (e.g., rotation, shear, shift) and uses an ensemble of adversarially trained models. R\&P~\cite{xie2018mitigating} introduces random resizing and padding, and Bit-Red~\cite{xu2018feature} reduces color depth and applies smoothing. FD~\cite{liu2019feature} proposes JPEG-based compression to neutralize perturbations while maintaining accuracy. NRP~\cite{naseer2020self} trains a self-supervised purifier network, and RS~\cite{cohen2019certified} adopts randomized smoothing to obtain provably $\ell_2$-robust classifiers.


\begin{figure*}[tb]
    \centering
    \begin{subfigure}{0.46\textwidth}
    \centering
    \begin{tikzpicture}
    \begin{customlegend}[legend columns=4,legend style={align=left,/tikz/every even column/.append style={column sep=0.2cm}},
            legend entries={{\tiny MIM}, 
                            {\tiny Freq. 40-299},
                            {\tiny Freq. 140-299},
                            {\tiny Freq. 240-299},
                            }]
            \addlegendimage{barlgend=Orange}
            \addlegendimage{barlgend=Green}
            
            \addlegendimage{barlgend=Blue}
            \addlegendimage{barlgend=Purple}
            \end{customlegend}
    \end{tikzpicture}
    \vspace{0.1em}
    \\
    \centering
    \pgfplotstableread[col sep=comma]{figs/data/high-freq-exp2.csv}\datatable
    \begin{tikzpicture}
        \begin{axis}[
            xtick pos=left,
            ybar,
            bar width=0.08cm,
            ylabel=\tiny Attack success rates (\%),
            width=0.95\linewidth,
            height=0.45\linewidth,
            xtick={0,1,2,3,4,5,6},
            xticklabels={\tiny VGG-16, Inc-v4, IncRes-v2, Res152-v2, Inc-v3$_{ens3}$, Inc-v3$_{ens4}$, IncRes-v2$_{ens}$},
            ymajorgrids=true,
            grid style=dashed,
            xticklabel style={rotate=15, font=\tiny},
            yticklabel style={font=\tiny},
            legend style={at={(0.95, 0.95)}, legend columns=-1, font=\tiny, /tikz/every even column/.append style={column sep=0.15cm}},
            legend image code/.code={
                \draw [#1, draw=none] (0cm,-0.1cm) rectangle (0.35cm,0.1cm); },
            ylabel style={font=\tiny, yshift=-5pt},
            xlabel style={font=\tiny, yshift=5pt},
            tick label style={font=\tiny},
            legend style={font=\tiny, fill opacity=0.8}
        ]
        
        \addplot[fill=Orange, draw=none, xshift=0.14cm] table[x expr=\coordindex, y=mifgsm] {\datatable};
        \addplot[fill=Green, draw=none, xshift=0.07cm] table[x expr=\coordindex, y=forty] {\datatable};
        \addplot[fill=Blue, draw=none,] table[x expr=\coordindex, y=H_forty] {\datatable};
        \addplot[fill=Purple, draw=none, xshift=-0.07cm] table[x expr=\coordindex, y=T_forty] {\datatable};
        \end{axis}
    \end{tikzpicture}
    \vspace{-1em}
    \caption{Attack success rates (\%) on seven black-box models with various sizes of perturbed region in the high-frequency space. The adversarial examples are generated on Inc-v3.}
    \label{fig:freq_mask_result}
    \end{subfigure}%
    \hspace{1.5em}
    \centering
    \begin{subfigure}{0.47\textwidth}
        \centering
        \begin{subfigure}{0.23\linewidth}
            \includegraphics[width=\linewidth]{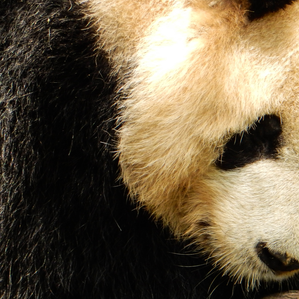}
            \caption*{Original Image}
        \end{subfigure}\hfill
        \begin{subfigure}{0.23\linewidth}
            \includegraphics[width=\linewidth]{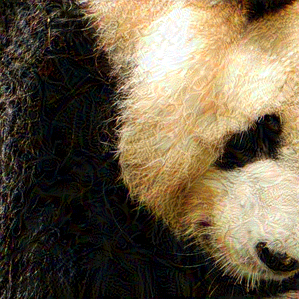}
            \caption*{Freq. 40-299}
        \end{subfigure}\hfill
        \begin{subfigure}{0.23\linewidth}
            \includegraphics[width=\linewidth]{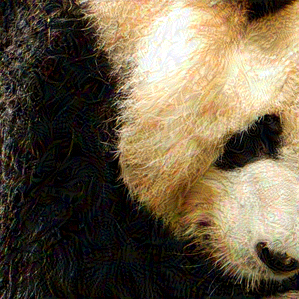}
            \caption*{Freq. 140-299}
        \end{subfigure}\hfill
        \begin{subfigure}{0.23\linewidth}
            \includegraphics[width=\linewidth]{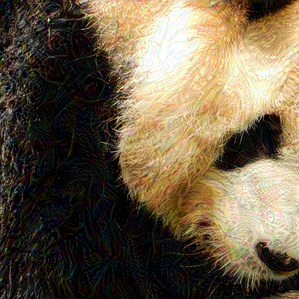}
            \caption*{Freq. 240-299}
        \end{subfigure}\hfill
        \caption{Raw image and the recovered images with various sizes of the perturbing region in the high-frequency space by IDCT.}
        \label{fig:freq_mask_visual}
    \end{subfigure}
    \vspace{-0.5em}
    \caption{Attack performance by perturbing various sizes of high-frequency components and the visualization of corresponding recovered images.}
    \label{fig:freq_mask}
    \vspace{-1em}
\end{figure*}

\subsection{Frequency-based Analysis of CNNs.}
Recent studies have explored neural network behaviors in the frequency domain. Wang~\etal~\cite{wang2020high} found that high-frequency components significantly enhance CNN accuracy. Yin~\etal~\cite{yin2019fourier} showed that models trained under natural conditions are more sensitive to high-frequency perturbations. Wang~\etal~\cite{wang2020frequencybased} revealed that a model's vulnerability to slight distortions is due to its reliance on high-frequency features, which attackers target to alter predictions. These findings suggest that disrupting high-frequency information could help identify crucial features more effectively, rather than exclusively focusing on low-frequency components as prior works have predominantly done.

\section{Method}
\subsection{Preliminaries}
Given a classifier $f(x;\theta)$ with parameters $\theta$, where $x$ is the benign image with the ground-truth label $y$. Adversarial attacks aim to add imperceptible perturbation to the clean image to craft an adversarial example $x^{adv}$, which satisfies $\Vert x^{adv} - x \Vert_{p} \leq \epsilon$, and mislead the classifier, \ie, $f(x^{adv};\theta) \neq y$. The process of generating adversarial examples can be formulated as an optimization problem:
\begin{equation}
    x^{adv} = \argmax_{x'}\ J(x',y;\theta),\ s.t.\  \Vert x' - x \Vert_{p} \leq \epsilon,
    \label{eq:opt_loss}
\end{equation}
where $J(\cdot)$ denotes the objective loss function (\eg, cross-entropy loss), and $\epsilon$ represents the upper bound of perturbation applied to the image. To ensure consistency with previous works~\citep{dong2018boosting, lin2020nesterov}, we focus on untargeted attacks and set $p = \infty$ in this work. 

\subsection{Motivation}
Since DNNs often share similar features~\citep{wu2020boosting}, feature-based attacks~\citep{zhou2018transferable,huang2019enhancing} aim to disrupt intermediate-level features to improve adversarial transferability. Existing feature-level attacks~\citep{wang2021feature,zhang2022enhancing} mainly perturb features using a weight matrix, which is computed on several transformed images to identify important features. We observe that these attacks mainly manipulate semantic information to figure out important features. From the perspective of the frequency domain, most of the semantic information in natural images is located in the low-frequency components of the spectrum~\citep{wallace1991jpeg, guo2018low}. However, many studies~\citep{wang2020high,wang2020frequencybased} have shown that CNNs tend to rely on high-frequency components for predictions, which are related to the abstract information in the image, \eg, textures and edges. Based on the above analysis, we notice that current feature-level attacks derive weight matrix mainly from semantic information, which might limit the adversarial transferability. It inspires us to explore the possibility of generating a weight matrix by jointly perturbing the semantic (low-frequency) and abstract (high-frequency) information of the image to enhance the adversarial transferability.

To validate the significance of high-frequency components in identifying the critical features for adversarial transferability, we add random noise to the high-frequency component to calculate the weight matrix, following the methodology proposed in FIA~\citep{wang2021feature}. Specifically, we progressively expand a square perturbation region from the lower-right corner of the frequency spectrum toward the upper-left, thereby covering a broader range of high-frequency bands. For an input image of size $299 \times 299 \times 3$, we define the high-frequency region as a square area of size $\tau \times \tau$ located at the lower-right corner of the frequency map, where $\tau \in \{40, 140, 240\}$. We adopt MIM as our backbone attack. The adversarial examples are generated on the Inc-v3 model and evaluated in the other seven models as outlined in Sec.~\ref{Exp_setup}.


Fig.~\ref{fig:freq_mask} (a) shows the experiment results. We can observe that perturbing the high-frequency region results in a higher attack success rate of attacking the seven models compared to the baseline approach. Notably, there is an observable enhancement in attack performance correlating with the expansion of the perturbed region. These findings corroborate the hypothesis that the disruption of high-frequency components, which encapsulate abstract information, plays a critical role in effectively pinpointing pivotal features that augment transferability. Furthermore, increasing the extent of frequency component perturbation contributes to more transferable adversarial examples.

\subsection{Frequency Domain Perturbation}
Based on the above analysis, the disruption of high-frequency components is beneficial in generating adversarial examples. High-frequency components are generally associated with abstract features, such as textures and edges, whereas low-frequency components correspond to semantic features. Consequently, perturbing the entire frequency spectrum facilitates the concurrent disturbance of both semantic and abstract features. It is noteworthy that the low-frequency components, especially the top-left elements in the frequency spectrum, play a crucial role in semantic features. Even little perturbation can significantly alter the semantic features, resulting in a recovered image that deviates substantially from the original ones. Inspired by the \textit{Admix} technique~\citep{wang2021admix}, we propose a novel method termed \textsc{Self-Mix}, which operates in the frequency domain to effectively disrupt discriminative features while preserving the semantic content of the image. Specifically, let $\mathcal{D}$ and $\mathcal{D}^{-1}$ denote the Discrete Cosine Transform (DCT) and its inverse (IDCT), respectively. Given an input image $x$, we first compute its frequency representation $\mathcal{D}(x)$ and then apply a random rotation by an angle $\beta$ in the spectral domain. The rotated spectrum is subsequently combined with the original spectrum to form a modified representation, which is then transformed back to the spatial domain using $\mathcal{D}^{-1}$. The process is mathematically represented as:
\begin{equation}
    \textsc{Self-Mix}(x, \mu, \beta) = \mathcal{D}^{-1}[\mathcal{D}(x) + \mu \cdot \mathcal{R}(\mathcal{D}(x), \beta)],
\end{equation}
where $\mu \in [0, 1]$ is the mixing strength and $\mathcal{R(\cdot)}$ indicates the rotation operation. Since the added spectrum is derived from the original spectrum, and the original spectrum has a dominant presence, this approach ensures that the image recovered through IDCT does not lose substantial content.

\subsection{Balancing the Semantic and Abstract Feature Disruption}
Although \textsc{Self-Mix} can perturb both abstract and semantic features, it does not perturb these two features with equal intensity. As shown in Fig.~\ref{fig:freq_mask} (b), disrupting high-frequency components has little effect on the image's semantic content. In the frequency domain, the proportion of low-frequency information is considerably smaller than high-frequency information. Moreover, the \textsc{Self-Mix} on the frequency spectrum induces minor perturbations in the semantic features. Thus, introducing equal-intensity perturbations directly across the frequency domain leads to an imbalance in the degree of perturbation for abstract and semantic features. This leads to a predominant focus on abstract features, while making semantic features receive less attention when identifying crucial features. In other words, the attacker disrupts more abstract features, leaving a greater portion of semantic features unaltered. This could render the meticulously designed adversarial examples less transferable. Therefore, it is crucial to strike a balance between the perturbation of these two kinds of features. 

To achieve this, we perform additional transformations on the benign image before applying \textsc{Self-Mix} to disrupt its semantic information. Specifically, we arbitrarily partition the original image into $n_b \times n_b$ blocks and randomly replace them with blocks from images of different categories. We refer to this operation as \textsc{BlockMix}, which is defined as follows: 
\begin{equation} {\small \textsc{BlockMix}(x,  \tilde{x})_{i,j} = \begin{cases}  \tilde{x}_{i,j}, &\text{with probability } 1 - p \\ x_{i,j}, &\text{with probability } p \end{cases},} \label{eq:BlockMix} \end{equation} 
where $\textsc{BlockMix}(x,  \tilde{x})_{i,j}$ denotes the resulting image block at position $(i, j)$, $x_{i,j}$ corresponds to the block from the original image $x$, and $ \tilde{x}_{i,j}$ corresponds to the block from an image $ \tilde{x}$ sampled from a different category.

Note that \textsc{BlockMix} primarily disrupts semantic information by altering image content. Subsequent \textsc{Self-Mix} perturb both high- and low-frequency components, thereby distorting abstract features along with certain semantic details. By combining these two approaches, a balance is achieved in terms of disrupting both abstract and semantic features, ensuring appropriate attention is paid to both when identifying key features.

\begin{algorithm}[t]
\caption{Semantic and Abstract Features Disruption}
\label{alg:algorithm}
\begin{flushleft}
\textbf{Input:} The original clean image $x$ with ground-truth label $y$, classification model $f$. \\
\textbf{Parameter:} Target layer $k$, keep probability $p$, the mixing strength $\mu$, angle of rotation $\beta$, ensemble number $N$, max perturbation $\epsilon$, the number of iteration $T$. \\
\textbf{Output:} The adversarial example $x_T^{\text{adv}}$.
\end{flushleft}
\vspace{0.5em}
\begin{algorithmic}[1]
    \State Let $\Delta=0$, $g_{0}=0$, $\mu=1$, $\alpha=\epsilon/T$
    \For{$n=0$ to $N-1$} \Comment{Calculating the weight matrix}
        \State Random sample an image $ \tilde{x}$ from other category
        \State $x_B = \textsc{BlockMix}(x,  \tilde{x})$
        \State $x_{SM} = \textsc{Self-Mix}(x_B, \mu, \beta)$
        \State $\Delta \gets \Delta + \frac{\partial J(x_{SM}, y;\theta)}{\partial f_k(x)}$
    \EndFor
    \State $\Delta \gets \Delta / \|\Delta\|_{2}$
    \State Construct objective: $L(x^{\text{adv}}) = \sum (\Delta \odot f_k(x^{\text{adv}}))$
    \State $x_0^{\text{adv}} = x$ \Comment{Update $x^{\text{adv}}$ by MIM}
    \For{$t=0$ to $T-1$}
        \State $g_{t+1} = \mu \cdot g_t + \frac{\nabla_x L(x_t^{\text{adv}})}{\|\nabla_x L(x_t^{\text{adv}})\|_{1}}$
        \State $x_{t+1}^{\text{adv}} = \text{Clip}_{x,\epsilon}(x_t^{\text{adv}} - \alpha \cdot \text{sign}(g_{t+1}))$
    \EndFor
    \State \Return $x_T^{\text{adv}}$
\end{algorithmic}
\end{algorithm}

\subsection{Attack Algorithm}
Based on the process of perturbation on semantic and abstract features, we refer to our proposed attack as Semantic and Abstract FEatures disRuption (SAFER). Specifically, we produce multiple transformed input images by applying \textsc{BlockMix} followed by \textsc{Self-Mix}. After that, we use the same strategy as FIA by calculating the average gradient \wrt the intermediate-level features of transformed images as the weight matrix to measure feature importance, which highlights consistently crucial semantic and abstract features for classification. 
Then, we utilize the sum of the element-wise product of the weight matrix and feature spectrum as the advanced objective function and adopt MIM as the backbone attack to guide the generation of adversarial example $x^{adv}$. The advanced objective function can be summarized as:
\begin{equation}
    L(x^{adv}) = \sum (\Delta \odot f_{k}(x^{adv})),
    \label{eq:loss_function}
\end{equation}
where $\Delta$ is the weight matrix and $f_{k}(x^{adv})$ is the $k$-th layer feature. By explicitly encouraging perturbations that simultaneously disrupt both abstract and semantic features, SAFER significantly enhances adversarial transferability. The overall procedure is detailed in Algorithm~\ref{alg:algorithm}.

\begin{table*}[tb]
    \centering
    \caption{Attack success rates (\%) of feature-level attacks in the single model setting. Adversarial examples are generated on Inc-v3, Inc-v4, Res-152, and VGG-16, respectively. * indicates the white-box model. The best results are in \textbf{bold}.}
    \resizebox{0.85\linewidth}{!}{\begin{tabular}{cc*{8}{b}}
        \toprule
        Model & Attack & Inc-v3 & Inc-v4 & IncRes-v2 & Res-152 & VGG-16 & Inc-v3$_\mathrm{ens3}$ & Inc-v3$_\mathrm{ens4}$ & IncRes-v2$_\mathrm{ens}$\\
        \midrule
        \multirow{5}{*}{Inc-v3} & MIM & \bf{100.0*} & 42.4 & 39.8 & 33.0 & 39.6 & 15.4 & 15.9 & ~~7.7 \\
        & FIA & ~~98.3* & 83.3 & 80.1 & 72.4 & 71.4 & 43.3 & 43.6 & 23.5 \\
        & RPA & ~~97.9* & 84.1 & 82.4 & 77.7 & 75.7 & 44.8 & 45.0 & 25.7 \\
        & NAA & ~~97.0* & 82.9 & 81.3 & 74.7 & 70.1 & 49.9 & 50.2 & 30.2 \\
        & \name & ~~97.8* & \setrow{\bfseries}87.2 & 85.9 & 78.6 & 78.2 & 50.5 & 51.2 & 31.5 \clearrow\\
        \midrule
        \multirow{5}{*}{Inc-v4} & MIM & 59.7 & \bf{100.0*} & 45.3 & 38.8 & 47.7 & 18.5 & 18.3 & ~~9.2 \\
        & FIA & 75.0 & ~~90.2* & 70.4 & 65.2 & 65.5 & 39.4 & 39.2 & 23.8 \\
        & RPA & 79.1 & ~~92.8* & 75.2 & 69.0 & 70.2 & 44.2 & 43.5 & 25.7 \\
        & NAA & 81.8 & ~~96.1* & 76.1 & 71.4 & 70.2 & 47.2 & 45.7 & 31.2 \\
        & \name & \setrow{\bfseries}85.7 \clearrow& ~~96.3* & \setrow{\bfseries}81.4 & 78.8 & 76.8 & 51.3 & 49.0 & 31.4 \clearrow\\
        \midrule
        \multirow{5}{*}{Res-152} & MIM & 52.6 & 47.8 & 44.9 & ~~\bf{99.5*} & 50.3 & 24.5 & 24.3 & 12.0 \\
        & FIA & 80.6 & 78.6 & 77.6 & ~~98.2* & 75.9 & 52.9 & 48.6 & 34.0 \\
        & RPA & 81.4 & 80.1 & 80.2 & ~~98.0* & 76.4 & 56.4 & 50.8 & 37.6 \\
        & NAA & 83.9 & 82.2 & 80.4 & ~~97.5* & 78.7 & 59.5 & 56.3 & 43.5 \\
        & \name & \setrow{\bfseries}87.1 & 86.0 & 84.5\clearrow & ~~98.8* & \setrow{\bfseries}83.3 & 61.5 & 58.6 & 44.9\clearrow \\
        \midrule
        \multirow{5}{*}{VGG-16} & MIM & 83.0 & 81.6 & 76.4 & 79.5 & \bf{100.0*} & 76.6 & 73.2 & 62.2 \\
        & FIA & 95.7 & 96.7 & 94.3 & 94.2 & \bf{100.0*} & 91.8 & 92.3 & 86.6 \\
        & RPA & 96.2 & 96.3 & 93.4 & 94.1 & \bf{100.0*} & 92.5 & 93.2 & 88.3 \\
        & NAA & 94.5 & 93.4 & 91.1 & 92.3 & ~~98.3* & 91.1 & 90.3 & 82.6 \\
        & \name & \setrow{\bfseries}97.2 & 97.5 & 95.5 & 94.7\clearrow & ~~99.9* & \setrow{\bfseries}93.5 & 93.4 & 90.0 \clearrow\\
        \bottomrule
    \end{tabular}}
    \label{tab:attack}
\end{table*}

\section{Experiments}
\subsection{Experiment Setup}
\label{Exp_setup}

\textbf{Dataset}. We adopt the ImageNet-compatible dataset for our experiments, which is widely used in previous works ~\citep{wang2021feature, zhang2022improving}. It contains 1,000 images with the size of $299 \times 299 \times 3$.

\textbf{Models}. To validate the effectiveness of \name, we adopt five widely adopted CNNs, namely, Inception-v3 (Inc-v3)~\citep{szegedy2016rethinking}, Inception-v4 (Inc-v4), Inception-Resnet-v2 (IncRes-v2)~\citep{szegedy2017inception}, Resnet-v2-152 (Res-152)~\citep{he2016deep}, VGG-16~\citep{simonyan2015very} and five vision transformers (ViT), \ie, PiT-B~\citep{heo2021rethinking}, CaiT-S~\citep{touvron2021going}, DeiT-B~\citep{touvron2021training}, Visformer-S~\citep{chen2021visformer}, Swin-T~\citep{liu2021swin}. To further verify its attack performance, we consider three ensemble adversarially trained models~\citep{kurakin2018ensemble}, \ie, Inc-v3$_\mathrm{ens3}$, Inc-v3$_\mathrm{ens4}$, IncRes-v2$_\mathrm{ens}$ and eight advanced defenses, \ie, HGD~\citep{liao2018defense}, R\&P~\citep{xie2018mitigating}, NIPS-r3, JPEG~\cite{karolina2016study}, Bit-Red~\citep{xu2018feature}, FD~\citep{liu2019feature}, RS~\citep{cohen2019certified}, NRP~\citep{naseer2020self}. 


\begin{table}[tb]
    \centering
    \caption{Attack success rates (\%) on five ViTs. The adversarial examples are generated on Inc-v3. 
    }
    \vspace{-.5em}
    \resizebox{\linewidth}{!}{
    \begin{tabular}{c*{5}{b}}
        \toprule
         Attack & PiT-B & CaiT-S & DeiT-B & Visformer-S & Swin-T \\
         \midrule
         MIM & 17.0 & 13.9 & 13.0 & 20.1 & 20.4 \\
         FIA & 37.4 & 37.8 & 34.6 & 50.4 & 50.9 \\
         RPA & 40.0 & 41.0 & 35.8 & 52.2 & 54.0 \\
         NAA & 44.2 & 42.1 & 39.4 & 53.6 & 54.3 \\
         \name & \setrow{\bfseries}46.8 & 46.0 & 43.8 & 59.5 & 59.0\clearrow \\
         \bottomrule
    \end{tabular}
    }
    \label{tab:attack_vit}
\end{table}
\begin{table}[tb]
    \centering
    \caption{Attack success rates (\%) on eight advanced defense methods. The adversarial examples are generated on Inc-v3. 
    }
    \vspace{-.5em}
    \resizebox{\linewidth}{!}{
    \begin{tabular}{c*{8}{b}}
    \toprule
    Attack & HGD & R\&P & NIPS-r3 & JPEG & Bit-Red & FD & RS & NRP \\
    \midrule
    MIM   & ~~5.1 & ~~7.9 & 10.7 & 29.6 & 23.4 & 39.5 & 28.9 & 2.5 \\
    FIA   & 10.5 & 19.3 & 28.0 & 68.0 & 41.9 & 62.3 & 32.7 & 2.8 \\
    RPA   & 16.6 & 26.6 & 37.2 & 72.5 & 46.0 & 62.4 & 33.1 & 5.3 \\
    NAA   & 22.0 & 33.6 & 43.0 & 70.6 & 44.2 & 58.8 & 33.6 & 5.9  \\
    \name  & \setrow{\bfseries}24.1 & 34.5 & 44.8 & 76.4 & 49.7 & 64.2 & 35.3 & 6.8\clearrow \\
    \bottomrule
    \end{tabular}
    }
    \label{tab:attack_defense}
\end{table}

\textbf{Baselines}. We adopt three feature-level attacks as our baselines with MIM~\citep{dong2018boosting} as the backbone attack, namely FIA~\citep{wang2021feature}, RPA~\citep{zhang2022enhancing}, and NAA~\citep{zhang2022improving}. We also integrate the baselines with three input transformation-based attacks (\ie, DIM~\citep{xie2019improving}, Admix~\citep{wang2021admix}, SSA~\citep{long2022frequency}) and three architecture-related attacks (\ie, SGM~\citep{Wu2020Skip}, LinBP~\citep{guo2020backpropagating}, BPA~\citep{wang2023rethinking}) to further validate the superiority of our method.

\textbf{Hyper-parameters}. We select the target layer in the source models same as in FIA~\citep{wang2021feature} and NAA~\citep{zhang2022improving}. We use a maximum perturbation of $\epsilon=16$, iterations $T=10$, and step size $\alpha=1.6$. For fairness, all feature-level attacks have a decay factor $\mu=1.0$ and ensemble number $N=30$. The drop probability $p_d$ in FIA and modification probability $p_m$ in RPA are set to 0.3 for standard models and adjusted to $p_d=0.1$ and $p_m=0.2$ for defense models and vision transformers. In RPA, the patch size is alternated as $n=1,3,5,7$. For NAA, $\gamma=1$. DIM uses a transform probability of 0.7. Admix sets admixed copies $m_1=5$, sampled images $m_2=3$, and admix strength $\eta=0.2$. SSA adopts tuning factor $\rho=0.5$, standard deviation $\sigma=16$, and the number of spectrum transformations 20. For architecture-related attacks, SGM has a decay factor of 0.5. We follow LinBP and BPA settings, modifying ReLU in the last eight residual blocks of ResNet-50. For our method, we set $n_b=5$, keep probability $p=0.9$, mixing strength coefficient $\mu=0.4$, and rotation angle $\beta \in [-\pi/4, \pi/4]$.
\subsection{Evaluations}

We evaluate the effectiveness of the proposed \name attack method with attack success rate, which denotes the ratio of all generated adversarial examples that can successfully mislead the target model.

\begin{figure*}[tb]
    \centering
    \begin{subfigure}{\linewidth}
    \centering
    \begin{tikzpicture}
    \begin{customlegend}[legend columns=5,legend style={align=left,/tikz/every even column/.append style={column sep=0.5cm}},
            legend entries={{\scriptsize Vanilla}, 
                            {\scriptsize +FIA},
                            {\scriptsize +RPA},
                            {\scriptsize +NAA},
                            {\scriptsize +\name},
                            }]
            \addlegendimage{barlgend=Orange}
            \addlegendimage{barlgend=Green}
            \addlegendimage{barlgend=Blue}
            \addlegendimage{barlgend=Purple}
            \addlegendimage{barlgend=Red}
    \end{customlegend}
    \end{tikzpicture}
    \vspace{0.7em}
    \end{subfigure}
    \\
    \begin{subfigure}{.33\textwidth}
    \centering
    \pgfplotstableread[col sep=comma]{figs/data/dim.csv}\datatable
    \begin{tikzpicture}
        \begin{axis}[
            xtick pos=left,
            ybar,
            bar width=0.08cm,
            ylabel={Attack success rates (\%)},
            width=\linewidth,
            height=0.7\linewidth,
            xtick=data,
            xticklabels from table={\datatable}{Category},
            ymajorgrids=true,
            grid style=dashed,
            xticklabel style={rotate=25, font=\tiny},
            yticklabel style={font=\tiny},
            legend style={at={(0.95, 0.95)}, legend columns=-1, font=\tiny, /tikz/every even column/.append style={column sep=0.15cm}},
            legend image code/.code={
                \draw [#1, draw=none] (0cm,-0.1cm) rectangle (0.35cm,0.1cm); },
            ylabel style={font=\tiny, yshift=-5pt},
            xlabel style={font=\tiny, yshift=5pt},
            tick label style={font=\tiny},
        ]
        
        \addplot[fill=Orange, draw=none, xshift=0.14cm] table[x expr=\coordindex, y=DIM] {\datatable};
        \addplot[fill=Green, draw=none, xshift=0.07cm] table[x expr=\coordindex, y=FIA] {\datatable};
        \addplot[fill=Blue, draw=none,] table[x expr=\coordindex, y=RPA] {\datatable};
        \addplot[fill=Purple, draw=none, xshift=-0.07cm] table[x expr=\coordindex, y=NAA] {\datatable};
        \addplot[fill=Red, draw=none, xshift=-0.14cm] table[x expr=\coordindex, y=Ours] {\datatable};
        
        
        \end{axis}
    \end{tikzpicture}
    \vspace{-1em}
    \caption{DIM}
    \end{subfigure}
    \begin{subfigure}{.33\textwidth}
    \centering
    \pgfplotstableread[col sep=comma]{figs/data/admix.csv}\datatable
    \begin{tikzpicture}
        \begin{axis}[
            xtick pos=left,
            ybar,
            bar width=0.08cm,
            ylabel={Attack success rates (\%)},
            width=\linewidth,
            height=0.7\linewidth,
            xtick =data,
            xticklabels from table={\datatable}{Category},
            ymajorgrids=true,
            grid style=dashed,
            xticklabel style={rotate=25, font=\tiny},
            yticklabel style={font=\tiny},
            legend style={at={(0.95, 0.95)}, legend columns=-1, font=\tiny, /tikz/every even column/.append style={column sep=0.15cm}},
            legend image code/.code={
                \draw [#1, draw=none] (0cm,-0.1cm) rectangle (0.35cm,0.1cm); },
            ylabel style={font=\tiny, yshift=-5pt},
            xlabel style={font=\tiny, yshift=5pt},
            tick label style={font=\tiny},
        ]
        
        \addplot[fill=Orange, draw=none, xshift=0.14cm] table[x expr=\coordindex, y=Admix] {\datatable};
        \addplot[fill=Green, draw=none, xshift=0.07cm] table[x expr=\coordindex, y=FIA] {\datatable};
        \addplot[fill=Blue, draw=none,] table[x expr=\coordindex, y=RPA] {\datatable};
        \addplot[fill=Purple, draw=none, xshift=-0.07cm] table[x expr=\coordindex, y=NAA] {\datatable};
        \addplot[fill=Red, draw=none, xshift=-0.14cm] table[x expr=\coordindex, y=Ours] {\datatable};
        
        
        \end{axis}
    \end{tikzpicture}
    \vspace{-1em}
    \caption{Admix}
    \end{subfigure}
    \begin{subfigure}{0.33\textwidth}
    \centering
    \pgfplotstableread[col sep=comma]{figs/data/ssa.csv}\datatable
    \begin{tikzpicture}
        \begin{axis}[
            xtick pos=left,
            ybar,
            bar width=0.08cm,
            ylabel={Attack success rates (\%)},
            width=\linewidth,
            height=0.7\linewidth,
            xtick =data,
            xticklabels from table={\datatable}{Category},
            ymajorgrids=true,
            grid style=dashed,
            xticklabel style={rotate=25, font=\tiny},
            yticklabel style={font=\tiny},
            legend style={at={(0.95, 0.95)}, legend columns=-1, font=\tiny, /tikz/every even column/.append style={column sep=0.15cm}},
            legend image code/.code={
                \draw [#1, draw=none] (0cm,-0.1cm) rectangle (0.35cm,0.1cm); },
            ylabel style={font=\tiny, yshift=-5pt},
            xlabel style={font=\tiny, yshift=5pt},
            tick label style={font=\tiny},
        ]
        
        \addplot[fill=Orange, draw=none, xshift=0.14cm] table[x expr=\coordindex, y=SSA] {\datatable};
        \addplot[fill=Green, draw=none, xshift=0.07cm] table[x expr=\coordindex, y=FIA] {\datatable};
        \addplot[fill=Blue, draw=none,] table[x expr=\coordindex, y=RPA] {\datatable};
        \addplot[fill=Purple, draw=none, xshift=-0.07cm] table[x expr=\coordindex, y=NAA] {\datatable};
        \addplot[fill=Red, draw=none, xshift=-0.14cm] table[x expr=\coordindex, y=Ours] {\datatable};
        
        
        \end{axis}
    \end{tikzpicture}
    \vspace{-1em}
    \caption{SSA}
    \end{subfigure}
    \vspace{-2em}
    \caption{Attack success rates (\%) of various attacks when combined with three input transformation-based attacks, \ie, DIM, Admix and SSA, respectively. The adversarial examples are generated on Inc-v3.}
    \label{fig:combine_trans}
\end{figure*}
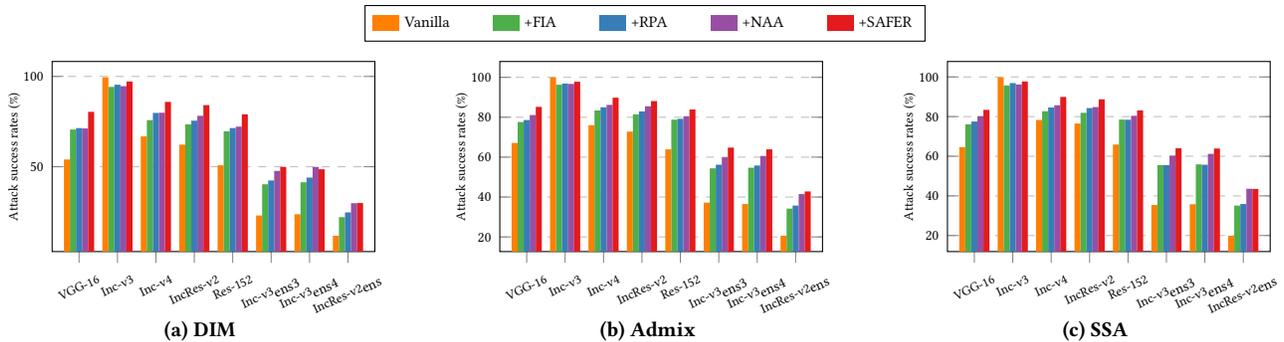
\begin{figure*}[tb]
    \centering
    \begin{subfigure}{\linewidth}
    \centering
    \begin{tikzpicture}
    \begin{customlegend}[legend columns=5,legend style={align=left,/tikz/every even column/.append style={column sep=0.5cm} },
            legend entries={{\scriptsize Vanilla}, 
                            {\scriptsize +FIA},
                            {\scriptsize +RPA},
                            {\scriptsize +NAA},
                            {\scriptsize +\name},
                            }]
            \addlegendimage{barlgend=Orange}
            \addlegendimage{barlgend=Green}
            \addlegendimage{barlgend=Blue}
            \addlegendimage{barlgend=Purple}
            \addlegendimage{barlgend=Red}
            \end{customlegend}
    \end{tikzpicture}
    \vspace{0.2em}
    \end{subfigure}
    \\
    \begin{subfigure}{.33\textwidth}
    \centering
    \pgfplotstableread[col sep=comma]{figs/data/sgm.csv}\datatable
    \begin{tikzpicture}
        \begin{axis}[
            xtick pos=left,
            ybar,
            bar width=0.08cm,
            ylabel={Attack success rates (\%)},
            width=\linewidth,
            height=0.7\linewidth,
            xtick =data,
            ymin=20,
            ymax=100,
            xticklabels from table={\datatable}{Category},
            ymajorgrids=true,
            grid style=dashed,
            xticklabel style={rotate=25, font=\tiny},
            yticklabel style={font=\tiny},
            legend style={at={(0.95, 0.95)}, legend columns=-1, font=\tiny, /tikz/every even column/.append style={column sep=0.15cm}},
            legend image code/.code={
                \draw [#1, draw=none] (0cm,-0.1cm) rectangle (0.35cm,0.1cm); },
            ylabel style={font=\tiny, yshift=-5pt},
            xlabel style={font=\tiny, yshift=5pt},
            tick label style={font=\tiny},
        ]
        
        \addplot[fill=Orange, draw=none, xshift=0.14cm] table[x expr=\coordindex, y=SGM] {\datatable};
        \addplot[fill=Green, draw=none, xshift=0.07cm] table[x expr=\coordindex, y=FIA] {\datatable};
        \addplot[fill=Blue, draw=none,] table[x expr=\coordindex, y=RPA] {\datatable};
        \addplot[fill=Purple, draw=none, xshift=-0.07cm] table[x expr=\coordindex, y=NAA] {\datatable};
        \addplot[fill=Red, draw=none, xshift=-0.14cm] table[x expr=\coordindex, y=Ours] {\datatable};
        \end{axis}
    \end{tikzpicture}
    \vspace{-1em}
    \caption{SGM}
    \end{subfigure}
    \begin{subfigure}{.33\textwidth}
    \centering
    \pgfplotstableread[col sep=comma]{figs/data/linbp.csv}\datatable
    \begin{tikzpicture}
        \begin{axis}[
            xtick pos=left,
            ybar,
            bar width=0.08cm,
            ylabel={Attack success rates (\%)},
            width=\linewidth,
            height=0.7\linewidth,
            xtick =data,
            ymin=20,
            ymax=100,
            xticklabels from table={\datatable}{Category},
            ymajorgrids=true,
            grid style=dashed,
            xticklabel style={rotate=25, font=\tiny},
            yticklabel style={font=\tiny},
            legend style={at={(0.95, 0.95)}, legend columns=-1, font=\tiny, /tikz/every even column/.append style={column sep=0.15cm}},
            legend image code/.code={
                \draw [#1, draw=none] (0cm,-0.1cm) rectangle (0.35cm,0.1cm); },
            ylabel style={font=\tiny, yshift=-5pt},
            xlabel style={font=\tiny, yshift=5pt},
            tick label style={font=\tiny},
        ]
        
        \addplot[fill=Orange, draw=none, xshift=0.14cm] table[x expr=\coordindex, y=LinBP] {\datatable};
        \addplot[fill=Green, draw=none, xshift=0.07cm] table[x expr=\coordindex, y=FIA] {\datatable};
        \addplot[fill=Blue, draw=none,] table[x expr=\coordindex, y=RPA] {\datatable};
        \addplot[fill=Purple, draw=none, xshift=-0.07cm] table[x expr=\coordindex, y=NAA] {\datatable};
        \addplot[fill=Red, draw=none, xshift=-0.14cm] table[x expr=\coordindex, y=Ours] {\datatable};
        \end{axis}
    \end{tikzpicture}
    \vspace{-1em}
    \caption{LinBP}
    \end{subfigure}
    \begin{subfigure}{.33\textwidth}
    \centering
    \pgfplotstableread[col sep=comma]{figs/data/bpa.csv}\datatable
    \begin{tikzpicture}
        \begin{axis}[
            xtick pos=left,
            ybar,
            bar width=0.08cm,
            ylabel={Attack success rates (\%)},
            width=\linewidth,
            height=0.7\linewidth,
            xtick =data,
            ymin=20,
            ymax=100,
            xticklabels from table={\datatable}{Category},
            ymajorgrids=true,
            grid style=dashed,
            xticklabel style={rotate=25, font=\tiny},
            yticklabel style={font=\tiny},
            legend style={at={(0.95, 0.95)}, legend columns=-1, font=\tiny, /tikz/every even column/.append style={column sep=0.15cm}},
            legend image code/.code={
                \draw [#1, draw=none] (0cm,-0.1cm) rectangle (0.35cm,0.1cm); },
            ylabel style={font=\tiny, yshift=-5pt},
            xlabel style={font=\tiny, yshift=5pt},
            tick label style={font=\tiny},
        ]
        
        \addplot[fill=Orange, draw=none, xshift=0.14cm] table[x expr=\coordindex, y=BPA] {\datatable};
        \addplot[fill=Green, draw=none, xshift=0.07cm] table[x expr=\coordindex, y=FIA] {\datatable};
        \addplot[fill=Blue, draw=none,] table[x expr=\coordindex, y=RPA] {\datatable};
        \addplot[fill=Purple, draw=none, xshift=-0.07cm] table[x expr=\coordindex, y=NAA] {\datatable};
        \addplot[fill=Red, draw=none, xshift=-0.14cm] table[x expr=\coordindex, y=Ours] {\datatable};
        \end{axis}
    \end{tikzpicture}
    \vspace{-1em}
    \caption{BPA}
    \end{subfigure}
    \vspace{-2em}
    \caption{Attack success rates (\%) of various attacks when combined with architecture-related attacks, \ie, SGM, LinBP and BPA, respectively. The adversarial examples are generated on Res-50.}
    \label{fig:combine_arch}
\end{figure*}
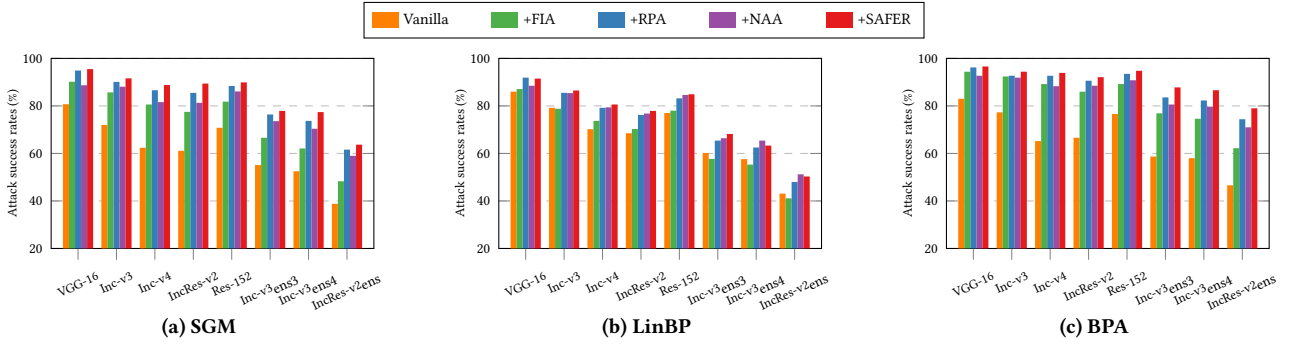
\textbf{Evaluation on feature-level attacks}. We first compare \name with various feature-level attacks and summarize the results in Tab.~\ref{tab:attack}. We can observe that feature-level attacks can significantly boost the adversarial transferability of MIM, and \name consistently outperforms the baselines on the black-box models. Compared with NAA which achieves the best attack performance among the baselines, \name outperforms NAA with an average margin of $3.1\%$ when generating the adversarial examples on Inc-v3. To further validate the effectiveness of our \name method, we also evaluate the attack success rates on five recently proposed vision transformer-based models. The adversarial examples are generated on Inc-v3 and the results are summarized in Tab.~\ref{tab:attack_vit}. \name maintains exceptional transferability, even when applied across vastly different architectural frameworks. In particular, \name attains the attack success rate of at least $43.8\%$ on five models, which outperforms the best baseline with a clear margin $2.6\% \sim 5.9\%$. This notable enhancement in performance on CNNs and ViTs affirms the outstanding effectiveness of our proposed method \name.

\textbf{Evaluation on input transformation-based attacks}. Among various types of transfer-based attacks, input transformation-based methods have gained significant attention. Our proposed method can be combined with these to improve adversarial example transferability. To validate \name's compatibility, we combined feature-level attacks with three input transformations, \ie, DIM, Admix, and SSA, using Inc-v3 as the source model, as shown in Fig.~\ref{fig:combine_trans}. The results show that while feature-level attacks may slightly reduce white-box performance, \name consistently achieves superior success rates compared to baselines. Regarding black-box transferability, \name outperforms existing feature-level attacks, surpassing the best baseline by margins of 3.9\%, 3.0\%, and 2.8\% for DIM, Admix, and SSA, respectively. These results underscore \name's superior compatibility with input transformation-based attacks.

\textbf{Evaluation on architecture-related attacks}. Recent studies have shown that modifying architecture-specific calculations in surrogate models can significantly enhance adversarial transferability. We further validate \name's compatibility with various architecture-related attacks in Fig.~\ref{fig:combine_arch}. Surprisingly, NAA, despite its prior superior performance, exhibits lower effectiveness when combined with these attacks, highlighting the need for comprehensive evaluations. In most cases, \name consistently achieves the best attack performance, outperforming the best baseline by an average margin of 2.1\%, 0.7\%, and 2.4\% for SGM, LinBP, and BPA, respectively. These results confirm the broad applicability and robust generality of \name across various attacks.

\begin{table*}[tb]
    \centering
    \caption{Attack success rates (\%) of various feature-level attacks in the ensemble model setting. The adversarial examples are generated on the ensemble models by combining Inc-v3, Inc-v4, and Res-152 individually. * indicates the white-box model. The best results are in \textbf{bold}.}
    \resizebox{0.85\linewidth}{!}{
    \begin{tabular}{cc*{7}{b}}
        \toprule
        Attack & Inc-v3 & Inc-v4 & IncRes-v2 & Res-152 & VGG-16 & Inc-v3$_\mathrm{ens3}$ & Inc-v3$_\mathrm{ens4}$ & IncRes-v2$_\mathrm{ens}$\\
        \midrule
        MIM & \textbf{99.9*} & 95.3* & 69.8 & 95.4* & 60.7 & 32.0 & 31.8 & 18.6 \\
        FIA & 94.6* & 88.8* & 83.7 & 97.8* & 79.4 & 60.9 & 57.5 & 42.1 \\
        RPA & 96.3* & 92.6* & 86.8 & 98.0* & 81.2 & 60.6 & 57.9 & 45.4 \\
        NAA & 96.0* & 94.7* & 87.6 & 97.8* & 82.0 & 66.0 & 65.2 & 51.2 \\
        \name & 97.4* & \setrow{\bfseries}95.7* & 90.5 & 98.9* & 88.6 & 71.8 & 69.7 & 53.8\clearrow \\
        \midrule
        MIM & \textbf{100.0*} & 95.5* & 58.4 & 44.4 & 50.3 & 20.4 & 23.0 & 11.5 \\
        FIA & ~~95.7* & 89.1* & 78.9 & 74.5 & 71.9 & 48.3 & 48.5 & 29.3\\
        RPA & ~~96.4* & 92.0* & 81.6 & 75.3 & 75.9 & 52.0 & 50.6 & 30.3 \\
        NAA & ~~97.0* & 95.1* & 84.6 & 78.5 & 77.6 & 56.3 & 58.0 & 39.3 \\
        \name & ~~97.8* & \setrow{\bfseries}96.0* & 89.6 & 83.8 & 84.4 & 58.6 & 60.1 & 39.5\clearrow \\
        \midrule
        MIM & \textbf{99.9*} & 61.3 & 59.5 & 95.2* & 53.5 & 26.6 & 26.8 & 14.3 \\
        FIA & 95.4* & 83.1 & 83.6 & 97.8* & 78.9 & 57.7 & 55.0 & 39.1 \\
        RPA & 96.1* & 84.9 & 85.2 & 98.4* & 80.3 & 58.5 & 56.3 & 40.7 \\
        NAA & 95.8* & 86.4 & 86.1 & 97.1* & 81.8 & 63.1 & 61.5 & 48.0 \\
        \name & 97.8* & \setrow{\bfseries}88.6 & 88.1 & 98.9* & 86.4 & 67.0 & 64.1 & 49.4\clearrow \\
        \midrule
        MIM & 70.9 & \textbf{99.3*} & 60.6 & 96.2* & 57.6 & 31.3 & 29.9 & 17.0 \\
        FIA & 83.2 & 89.1* & 79.9 & 97.9* & 77.5 & 53.4 & 51.8 & 36.8 \\
        RPA & 85.7 & 92.6* & 82.3 & 98.0* & 79.4 & 57.4 & 54.6 & 41.5 \\
        NAA & 86.5 & 94.3* & 84.1 & 97.7* & 80.6 & 62.1 & 60.1 & 46.4 \\
        \name & \textbf{89.9} & 95.9* & \setrow{\bfseries}87.6 & 99.0* & 87.0 & 67.1 & 63.6 & 50.0\clearrow \\
        \bottomrule
    \end{tabular}}
    \label{tab:attack_ens}
\end{table*}

\textbf{Evaluation on ensemble models.}
Liu~\etal~\cite{liu2017delving} showed that attacking multiple models simultaneously enhances adversarial transferability. Building on this insight, we evaluate our proposed method under an ensemble-based attack setting to assess its effectiveness. For MIM, we followed the same ensemble scheme in its paper to fuse the logits. For feature-level attack methods, we took another ensemble scheme, which averaged the loss of each model. We mainly include three models in our study, which are Inc-v3, Inc-v4, and Res-152. In the experiments, we keep one model as the hold-out black-box model and attack an ensemble of the other two models. We also test the success rate of the attack on the normally and adversarially trained black-box models. The results are shown in Tab.~\ref{tab:attack_ens}. The adversarial examples generated by our proposed \name have higher attack success rates compared to other baseline methods. \name can outperform the baseline method by an average of 3.2\%, 3.5\%, 2.5\%, and 3.4\% for ensembling all three models or holding out one of them, respectively. The results suggest that \name can generate more transferable adversarial examples under the ensemble setting, which further validates its superiority.

\textbf{Evaluation on advanced defense methods}. To further validate the effectiveness of \name, we evaluated adversarial examples generated on Inc-v3 against eight powerful defense mechanisms. As shown in Tab.~\ref{tab:attack_defense}, the attack performance significantly decays, validating the effectiveness of these defense methods. Among the baselines, RPA performs best on JPEG, Bit-Red, and FD, while NAA excels on the other five defenses. In contrast, \name consistently outperforms both RPA and NAA across all eight defenses. 
This superior performance underscores its versatility and effectiveness in overcoming diverse advanced defenses.

In summary, \name demonstrates superior performance over existing feature-level attack methods across both CNNs and ViTs. It exhibits strong compatibility with a variety of input transformation-based and architecture-specific attack strategies, and consistently achieves higher attack success rates compared to baseline methods, even in the presence of advanced defense mechanisms. Furthermore, \name achieves notably strong results under ensemble attack settings, highlighting its effectiveness and generalizability in real-world adversarial scenarios.

\begin{table*}[tb]
    \centering
    \caption{Attack success rates (\%) of adversarial examples generated on Inc-v3 w/wo \textsc{BlockMix} or \textsc{Self-Mix} or both. * indicates the white-box model. The best results are in \textbf{bold}.
    }
    \resizebox{0.9\linewidth}{!}{
    \small
    \begin{tabular}{cc>{\rowmac}c>{\rowmac}c>{\rowmac}c>{\rowmac}c>{\rowmac}c>{\rowmac}c>{\rowmac}c>{\rowmac}c>{\rowmac}c>{\rowmac}c}
        \toprule
         \textsc{BlockMix} & \textsc{Self-Mix} & VGG-16 & Inc-v3 & Inc-v4 & IncRes-v2 & Res-152 & Inc-v3$_\mathrm{ens3}$ & Inc-v3$_\mathrm{ens4}$ & IncRes-v2$_\mathrm{ens}$\\
         \midrule
         \xmark & \xmark & 38.9 & \textbf{100.0*} & 41.8 & 38.3 & 32.6 & 14.8 & 15.3 & ~~7.6 \\
         \cmark & \xmark & 61.6 & 96.2* & 69.7 & 67.6 & 59.5 & 32.6 & 31.7 & 17.2 \\
         \xmark & \cmark & 70.9 & 96.0* & 78.9 & 77.7 & 72.0 & 47.9 & 48.5 & 30.0 \\
         \cmark & \cmark & \setrow{\bfseries}77.5 \clearrow & 98.5* & \setrow{\bfseries}86.5 & 85.0 & 78.2 & 52.5 & 52.1 & 31.2\clearrow \\
         \bottomrule
    \end{tabular}
    }
    \label{tab:ablation_method}
\end{table*}

\subsection{Ablation Study}
We conduct ablation and parameter studies to thoroughly understand \name, using Inc-v3 with MIM as the backbone method for generating adversarial examples.

\textbf{Abstract \vs semantic feature disruption}. We conducted ablation experiments to validate the importance of balancing disruptions to semantic and abstract features, summarized in Tab.~\ref{tab:ablation_method}. When applying only \textsc{BlockMix} or \textsc{Self-Mix}, the white-box attack success rate decreases slightly compared to the baseline. However, under the black-box setting, transferability significantly improves, with increases of 16.3\% and 32.6\% respectively over MIM. This validates that each transformation method effectively identifies and perturbs critical features, resulting in enhancing adversarial transferability. Moreover, the substantial improvement from perturbing both high and low-frequency components underscores the effectiveness of disrupting both abstract and semantic features to uncover crucial characteristics in our approach. Integrating both \textsc{BlockMix} and \textsc{Self-Mix} results in optimal attack performance, highlighting that uniform-intensity perturbations across the frequency spectrum disproportionately emphasize semantic features. It also highlights the significance and effectiveness of our balanced methodology, which disrupts both abstract and semantic features to calculate the weight matrix for better adversarial transferability.

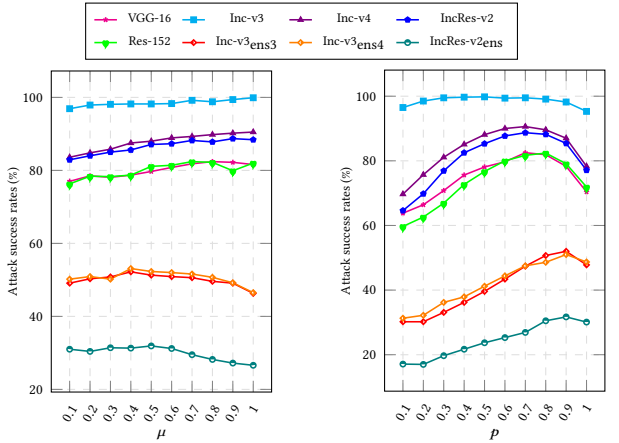
\begin{figure}[tb]
    \begin{subfigure}{\linewidth}
    \centering
    \begin{tikzpicture}
    \begin{customlegend}[legend columns=4,legend style={align=left,column sep=0.5ex},
    legend image post style={scale=0.5},
            legend entries={{\tiny VGG-16},
                            {\tiny Inc-v3}, 
                            {\tiny Inc-v4},
                            {\tiny IncRes-v2},
                            {\tiny Res-152},
                            {\tiny Inc-v3$_\mathrm{ens3}$},
                            {\tiny Inc-v3$_\mathrm{ens4}$},
                            {\tiny IncRes-v2$_\mathrm{ens}$},
                            }]
            \addlegendimage{line width=0.6pt, mark=star, solid, color=magenta, mark options={mark size=2pt}}  
            \addlegendimage{line width=0.6pt, mark=square*, solid, color=cyan, mark options={mark size=2pt}}
            \addlegendimage{line width=0.6pt, mark=triangle*, solid, color=violet, mark options={mark size=2pt}}
            \addlegendimage{line width=0.6pt, mark=pentagon*, solid, color=blue, mark options={mark size=2pt}}
            
            \addlegendimage{line width=0.6pt, mark=heart, solid, color=green, mark options={mark size=2pt}}
            \addlegendimage{line width=0.6pt, mark=halfsquare left*, solid, color=red, mark options={mark size=2pt}}
            \addlegendimage{line width=0.6pt, mark=halfsquare right*, solid, color=orange, mark options={mark size=2pt}}
            \addlegendimage{line width=0.6pt, mark=halfcircle*, solid, color=teal, mark options={mark size=2pt}}
            \end{customlegend}
    \end{tikzpicture}
    \vspace{0.3em}
    \end{subfigure}
    \begin{subfigure}{.48\linewidth}
    \centering
    \begin{tikzpicture}[clip]
        \begin{axis}[
        	xlabel= $\mu$, 
        	ylabel=Attack success rates (\%),
        	grid=both,
        	minor grid style={gray!25, dashed},
        	major grid style={gray!25, dashed},
            scale only axis,
            width=0.72\linewidth,
            height=1.05\linewidth,
            ylabel style={font=\tiny, yshift=-5pt},
            xlabel style={font=\tiny, yshift=5pt},
            legend pos=south east,
            legend style={align=center,font=\tiny,minimum width=0.1cm},
            xticklabel style={rotate=60},
            tick label style={font=\tiny},
            xtick distance=0.1,
        ]
            \addplot[line width=0.6pt,solid,mark=star,color=magenta, mark options={mark size=1pt}] %
            	table[x=mu,y=vgg16,col sep=comma]{figs/data/mu.csv};
            \addplot[line width=0.6pt,solid,mark=square*,color=cyan, mark options={mark size=1pt}] %
            	table[x=mu,y=inceptionv3,col sep=comma]{figs/data/mu.csv};
            \addplot[line width=0.6pt,solid,mark=triangle*,color=violet, mark options={mark size=1pt}] %
            	table[x=mu,y=inceptionv4,col sep=comma]{figs/data/mu.csv};
            \addplot[line width=0.6pt,solid,mark=pentagon*,color=blue, mark options={mark size=1pt}] %
            	table[x=mu,y=inception_resnet_v2,col sep=comma]{figs/data/mu.csv};
            \addplot[line width=0.6pt,solid,mark=heart,color=green, mark options={mark size=1pt}] %
            	table[x=mu,y=resnet_v2_152,col sep=comma]{figs/data/mu.csv};
            \addplot[line width=0.6pt,solid,mark=halfsquare left*,color=red, mark options={mark size=1pt}] %
            	table[x=mu,y=ens3_adv_inception_v3,col sep=comma]{figs/data/mu.csv};
            \addplot[line width=0.6pt,solid,mark=halfsquare right*,color=orange, mark options={mark size=1pt}] %
            	table[x=mu,y=ens4_adv_inception_v3,col sep=comma]{figs/data/mu.csv};
            \addplot[line width=0.6pt,solid,mark=halfcircle*,color=teal, mark options={mark size=1pt}] %
            	table[x=mu,y=ens_adv_inception_resnet_v2,col sep=comma]{figs/data/mu.csv};
        \end{axis}
        \end{tikzpicture}
        \vspace{-2mm}
    \caption{Attack success rates (\%) with various mixing strength $\mu$.}
    \label{fig:ablation:mu}
    \end{subfigure}%
    \hfill
    \hspace{0.2em}
    \begin{subfigure}{.48\linewidth}
    \centering
    \hspace{-0.1em}
    \begin{tikzpicture}[clip]
        \begin{axis}[
        	xlabel= $p$, 
        	ylabel=Attack success rates (\%),
        	grid=both,
        	minor grid style={gray!25, dashed},
        	major grid style={gray!25, dashed},
            scale only axis,
            width=0.72\linewidth,
            height=1.05\linewidth,
            ylabel style={font=\tiny, yshift=-5pt},
            xlabel style={font=\tiny, yshift=5pt},
            legend pos=south east,
            legend style={align=center,font=\tiny,minimum width=0.1cm},
            xticklabel style={rotate=60},
            tick label style={font=\tiny},
            xtick distance=0.1,
        ]
            \addplot[line width=0.6pt,solid,mark=star,color=magenta, mark options={mark size=1pt}] %
            	table[x=probb,y=vgg16,col sep=comma]{figs/data/probb.csv};
            \addplot[line width=0.6pt,solid,mark=square*,color=cyan, mark options={mark size=1pt}] %
            	table[x=probb,y=inceptionv3,col sep=comma]{figs/data/probb.csv};
            \addplot[line width=0.6pt,solid,mark=triangle*,color=violet, mark options={mark size=1pt}] %
            	table[x=probb,y=inceptionv4,col sep=comma]{figs/data/probb.csv};
            \addplot[line width=0.6pt,solid,mark=pentagon*,color=blue, mark options={mark size=1pt}] %
            	table[x=probb,y=inception_resnet_v2,col sep=comma]{figs/data/probb.csv};
            \addplot[line width=0.6pt,solid,mark=heart,color=green, mark options={mark size=1pt}] %
            	table[x=probb,y=resnet_v2_152,col sep=comma]{figs/data/probb.csv};
            \addplot[line width=0.6pt,solid,mark=halfsquare left*,color=red, mark options={mark size=1pt}] %
            	table[x=probb,y=ens3_adv_inception_v3,col sep=comma]{figs/data/probb.csv};
            \addplot[line width=0.6pt,solid,mark=halfsquare right*,color=orange, mark options={mark size=1pt}] %
            	table[x=probb,y=ens4_adv_inception_v3,col sep=comma]{figs/data/probb.csv};
            \addplot[line width=0.6pt,solid,mark=halfcircle*,color=teal, mark options={mark size=1pt}] %
            	table[x=probb,y=ens_adv_inception_resnet_v2,col sep=comma]{figs/data/probb.csv};
        \end{axis}
        \end{tikzpicture}
        \vspace{-2mm}
    \caption{Attack success rates (\%) with various keep probability $p$.}
    \label{fig:ablation:probb}
    \end{subfigure}%
    \vspace{-2mm}
    \caption{Hyper-parameter studies of \name on the mixing strength $\mu$ and keep probability $p$. The adversarial examples are generated on Inc-v3}
    \label{fig:ablation_peremeter}
\end{figure}
\textbf{The mixing strength $\mu$}. A critical aspect of implementing \textsc{Self-Mix} in the frequency spectrum is regulating the mixing strength to ensure the original frequency spectrum remains predominant. Parameter studies on mixing strength, shown in Fig.~\ref{fig:ablation_peremeter} (a), indicate that increasing the mixing strength enhances transferability, with an optimum at $\mu=0.4$ for ensemble adversarially trained models. While higher $\mu$ values improve performance on normally trained models, balancing performance across both normally and adversarially trained models is essential. Therefore, we adopt $\mu=0.4$.

\textbf{The keep probability $p$}. The keep probability plays a crucial role as it determines the degree to which semantic information in an image is preserved or altered.  Fig.~\ref{fig:ablation_peremeter} (b) illustrates the effect of altering the probability. We can observe that with the increase of the keep probability, the attack success rate on both normally or adversarially trained models increases. When continually increasing $p$, the attack success rates begin to slightly decrease when the probability is around 0.8 or 0.9. To ensure high attack performance, we set the probability $p=0.9$.

\begin{figure}[tb]
    \begin{subfigure}{\linewidth}
    \centering
    \begin{tikzpicture}
    \begin{customlegend}[legend columns=4,legend style={align=left,column sep=0.5ex},
    legend image post style={scale=0.5},
            legend entries={{\tiny VGG-16},
                            {\tiny Inc-v3}, 
                            {\tiny Inc-v4},
                            {\tiny IncRes-v2},
                            {\tiny Res-152},
                            {\tiny Inc-v3$_\mathrm{ens3}$},
                            {\tiny Inc-v3$_\mathrm{ens4}$},
                            {\tiny IncRes-v2$_\mathrm{ens}$},
                            }]
            \addlegendimage{line width=0.6pt, mark=star, solid, color=magenta, mark options={mark size=2pt}}  
            \addlegendimage{line width=0.6pt, mark=square*, solid, color=cyan, mark options={mark size=2pt}}
            \addlegendimage{line width=0.6pt, mark=triangle*, solid, color=purple, mark options={mark size=2pt}}
            \addlegendimage{line width=0.6pt, mark=pentagon*, solid, color=blue, mark options={mark size=2pt}}
            \addlegendimage{line width=0.6pt, mark=heart, solid, color=green, mark options={mark size=2pt}}
            \addlegendimage{line width=0.6pt, mark=halfsquare left*, solid, color=red, mark options={mark size=2pt}}
            \addlegendimage{line width=0.6pt, mark=halfsquare right*, solid, color=orange, mark options={mark size=2pt}}
            \addlegendimage{line width=0.6pt, mark=halfcircle*, solid, color=brown, mark options={mark size=2pt}}
            \end{customlegend}
    \end{tikzpicture}
    \vspace{0.3em}
    \end{subfigure}
    \begin{subfigure}{\linewidth}
    \centering
    \begin{tikzpicture}[clip]
        \begin{axis}[
        	xlabel= $k$, 
        	ylabel=Attack success rates (\%),
        	grid=both,
        	minor grid style={gray!25, dashed},
        	major grid style={gray!25, dashed},
            scale only axis,
            width=0.8\linewidth,
            height=0.4\linewidth,
            ylabel style={font=\scriptsize, yshift=-5pt},
            xlabel style={font=\scriptsize, yshift=5pt},
            xtick={0,1,2,3,4,5,6,7,8,9},
            xticklabels={Conv1a, Conv2b, Conv3b, Conv4a, Mix5b, Mix6a, Mix7a},
            legend pos=south east,
            legend style={align=center,font=\tiny,minimum width=0.1cm},
            xticklabel style={rotate=30},
            tick label style={font=\tiny},
            xtick distance=0.1,
        ]
            \addplot[line width=0.6pt,solid,mark=star,color=magenta, mark options={mark size=1pt}] %
            	table[x=target_layer,y=vgg16,col sep=comma]{figs/data/target_layer.csv};
            \addplot[line width=0.6pt,solid,mark=square*,color=cyan, mark options={mark size=1pt}] %
            	table[x=target_layer,y=inceptionv3,col sep=comma]{figs/data/target_layer.csv};
            \addplot[line width=0.6pt,solid,mark=triangle*,color=purple, mark options={mark size=1pt}] %
            	table[x=target_layer,y=inceptionv4,col sep=comma]{figs/data/target_layer.csv};
            \addplot[line width=0.6pt,solid,mark=pentagon*,color=blue, mark options={mark size=1pt}] %
            	table[x=target_layer,y=inception_resnet_v2,col sep=comma]{figs/data/target_layer.csv};
            \addplot[line width=0.6pt,solid,mark=heart,color=green, mark options={mark size=1pt}] %
            	table[x=target_layer,y=resnet_v2_152,col sep=comma]{figs/data/target_layer.csv};
            \addplot[line width=0.6pt,solid,mark=halfsquare left*,color=red, mark options={mark size=1pt}] %
            	table[x=target_layer,y=ens3_adv_inception_v3,col sep=comma]{figs/data/target_layer.csv};
            \addplot[line width=0.6pt,solid,mark=halfsquare right*,color=orange, mark options={mark size=1pt}] %
            	table[x=target_layer,y=ens4_adv_inception_v3,col sep=comma]{figs/data/target_layer.csv};
            \addplot[line width=0.6pt,solid,mark=halfcircle*,color=brown, mark options={mark size=1pt}] %
            	table[x=target_layer,y=ens_adv_inception_resnet_v2,col sep=comma]{figs/data/target_layer.csv};
        \end{axis}
        \end{tikzpicture}
    \label{fig:ablation:target_layer}
    \end{subfigure}
    \vspace{-2em}
    \caption{Hyper-parameter studies of \name on the different target layer $k$. The adversarial examples are generated on Inc-v3}
    \label{fig:ablation_target_layer}
    \vspace{-1em}
\end{figure}
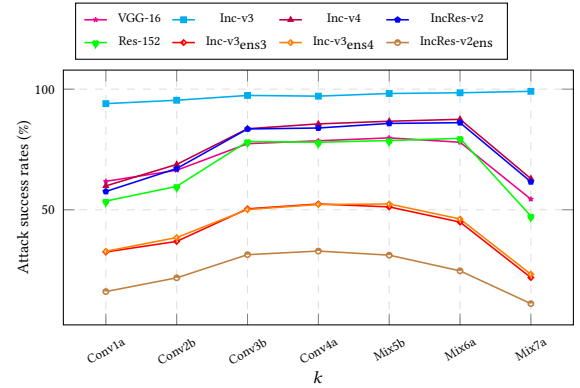
\textbf{The target layer $k$.}
The selection of the target layer has a very strong impact on feature-level attacks. Therefore, we conduct ablation experiments to investigate the impact of the target layer selection on the attack success rate. We choose the Inc-v3 as the source model. Fig.~\ref{fig:ablation_target_layer} displays the effect of feature layer selection on the attack success rate from shallow to deep. We can see that as the depth of the selected intermediate layer gradually deepens, the attack success rates of the attack on both the normally and the adversarially trained models increase. However, when the target layer becomes deeper and exceeds the Conv4a layer, the attack success rates on the adversarially trained models start to decrease but only marginally. When the target layer $k$ is deeper than the Mix6a layer, the attack success rates on both the normally and adversarially trained models show a significant decrease. In order to balance the performance on both types of models, we selected Mix5b as the target layer for Inc-v3, which is consistent in FIA.

\section{Conclusion}
In this work, we emphasize that current feature-level attacks typically focus on disrupting semantic information to calculate the weight matrix for feature importance, often neglecting the perturbation of abstract features associated with high-frequency information. Existing researches indicate that deep neural networks heavily rely on high-frequency components for accurate recognition. Building on this, we propose \fname, which aims to perturb both abstract and semantic features for better transferability. Specifically, our approach uses \textsc{BlockMix} to disrupt semantic content and \textsc{Self-Mix} on the frequency spectrum to perturb abstract features. Experimental results demonstrate that \name generates adversarial examples with superior transferability, outperforming baseline methods.

\bibliographystyle{ACM-Reference-Format}
\bibliography{sample-base}

\newpage
\clearpage
\appendix

\section{Further Evaluations}
In this section, we provide more experimental results to evaluate the proposed \name comprehensively. The experiments maintain the same experimental setup as in the main body of the paper.

\textbf{Evaluation on ViT models.}
To better validate the superiority of our approach, in addition to attacking five different ViT models using the adversarial samples generated on the Inc-v3 model, we also evaluate the success rates of the attacks on the adversarial samples generated on Inc-v4, Res-152, and VGG-16, respectively. As shown in Tab.~\ref{tab:attack_vit_sup}, we can observe that our proposed \name can generate more transferable adversarial samples on different source models. On average, \name outperforms the best baseline with a clear margin of 2.5\%, 3.5\%, and 3.7\% on Inc-v4, Res-152, and VGG-16, respectively. When using Res-152 as the source model and Swin-T as the target model, our proposed method is able to outperform the best baseline method in terms of attack success rate by 5.4\%. These results can be more evidence of the superiority of our proposed method.
\begin{table}[hb]
    \centering
    \caption{Attack success rates (\%) on five popular ViT models. The adversarial examples are generated on Inc-v4, Resnet-152, and VGG-16, respectively. The best results are in \textbf{bold}.}
    \resizebox{\linewidth}{!}{
    \begin{tabular}{cc*{5}{b}}
        \toprule
        Model & Attack & PiT-B & CaiT-S & DeiT-B & Visformer-S & Swin-T \\
        \midrule
        \multirow{5}{*}{Inc-v4} & MIM & 20.2 & 19.0 & 16.0 & 25.0 & 27.0 \\
        & FIA & 40.0 & 38.2 & 31.2 & 54.8 & 51.0 \\
        & RPA & 46.5 & 47.2 & 40.5 & 60.7 & 57.5 \\
        & NAA & 48.8 & 48.0 & 43.2 & 57.5 & 57.8 \\
        & \name& \setrow{\bfseries}51.2 & 50.2 & 44.8 & 61.5 & 60.2 \clearrow \\
        \midrule
        \multirow{5}{*}{Res-152} & MIM & 20.8 & 20.0 & 18.8 & 29.2 & 27.7 \\
        & FIA & 35.5 & 37.0 & 33.8 & 54.2 & 48.8 \\
        & RPA & 45.2 & 42.0 & 41.5 & 60.7 & 56.5 \\
        & NAA & 46.8 & 48.8 & 46.0 & 61.3 & 58.8 \\
        & \name & \setrow{\bfseries}50.8 & 51.5 & 47.8 & 65.0 & 64.2 \clearrow \\
        \midrule
        \multirow{5}{*}{VGG-16} & MIM & 47.8 & 54.2 & 54.0 & 65.5 & 60.0 \\
        & FIA & 57.2 & 69.8 & 68.5 & 84.8 & 69.0 \\
        & RPA & 70.2 & 78.8 & 77.5 & 92.5 & 80.2 \\
        & NAA & 74.2 & 76.5 & 76.5 & 87.8 & 83.2 \\
        & \name & \setrow{\bfseries}74.5 & 83.5 & 82.5 & 93.2 & 83.2 \clearrow \\
        \bottomrule
    \end{tabular}}
    \label{tab:attack_vit_sup}
\end{table}

\textbf{Evaluation on defense methods.}
Following the same experimental setup in the main body of the paper using the eight defense mechanisms, we further report the attack success rate of the adversarial examples generated on three additional source models including Inc-v4, Res-152, and VGG-16 in Tab.~\ref{tab:attack_defense_sup}. Our proposed \name achieves the highest attack success rates among all the tested methods when attacking eight different defense methods. Compared with the best results obtained from four baseline methods, there is a significant margin between the results of our method and the best ones. On Inc-v4, Res-152, and VGG-16, our method is able to outperform the best results among the baseline method by 2.6\%, 2.8\%, and 3.3\% on average, respectively. Specifically, \name outperforms the best baseline method with an obvious margin from 0.5\% to 7.0\% when the adversarial examples are generated on the Res-152. The experimental results fully demonstrate that our proposed method can pose an effective threat to existing defense methods.
\begin{table}[htb]
    \centering
    \caption{Attack success rates (\%) on eight advanced defense methods. The adversarial examples are generated on Inc-v4, Res-152, and VGG-16, respectively. Best results are in \textbf{bold}.}
    \resizebox{\linewidth}{!}{
    \begin{tabular}{cc*{8}{b}}
        \toprule
        Model & Attack & HGD & R\&P & NIPS-r3 & JPEG & Bit-Red & FD & RS & NRP \\
        \midrule
        \multirow{5}{*}{Inc-v4} & MIM & ~~7.1 & 10.7 & 13.5 & 30.2 & 25.9 & 42.6 & 28.3 & ~~6.8 \\
        & FIA & 17.2 & 21.2 & 30.2 & 69.8 & 52.3 & 64.2 & 32.3 & ~~7.2 \\
        & RPA & 23.8 & 25.8 & 36.8 & 73.9 & 57.6 & 65.4 & 31.9 & 12.7 \\
        & NAA & 29.2 & 32.0 & 40.1 & 74.6 & 57.9 & 66.8 & 31.9 & 17.2 \\
        & \name & \setrow{\bfseries}30.1 & 34.8 & 43.0 & 79.8 & 61.4 & 68.7 & 33.4 & 19.3\clearrow \\
        \midrule
        \multirow{5}{*}{Res-152} & MIM &15.8 & 14.7 & 16.7 & 37.6 & 30.5 & 47.6 & 29.6 & 10.3 \\
        & FIA & 34.5 & 30.5 & 40.6 & 76.2 & 56.1 & 70.6 & 35.4 & 15.6 \\
        & RPA & 42.3 & 38.1 & 49.4 & 79.6 & 59.4 & 76.4 & 35.5 & 19.8 \\
        & NAA & 49.4 & 45.8 & 53.1 & 80.3 & 60.3 & 78.3 & 35.8 & 21.3 \\
        & \name & \setrow{\bfseries}50.9 & 46.3 & 54.7 & 84.6 & 67.3 & 81.9 & 36.5 & 24.0\clearrow \\
        \midrule
        \multirow{5}{*}{VGG-16} & MIM & 75.2 & 51.9 & 80.3 & 88.6 & 82.3 & 85.4 & 34.8 & 28.9 \\
        & FIA & 89.3 & 72.3 & 89.9 & 90.4 & 86.6 & 86.8 & 37.9 & 36.4 \\
        & RPA & 92.8 & 80.9 & 93.6 & 94.3 & 89.7 & 89.1 & 43.5 & 38.7 \\
        & NAA & 89.6 & 76.8 & 92.6 & 93.4 & 87.8 & 90 & 42.3 & 37.1 \\
        & \name & \setrow{\bfseries}93.6 & 84.8 & 98.2 & 98.7 & 92.3 & 93.4 & 48.1 & 39.4\clearrow \\
        \bottomrule
    \end{tabular}}
    \label{tab:attack_defense_sup}
\end{table}

\textbf{Evaluation on gradient-based attacks}
Gradient-based attacks are an essential part of transfer-based attacks and have received extensive research and attention. To further validate the compatibility of our proposed method, we select three representative gradient-based methods, \ie, EMI~\citep{wang2021boosting}, GMI~\citep{wang2022boosting}, and VMI~\citep{wang2021enhancing}, as the backbone attacks integrated into the feature-level attacks. We follow the default settings in these works and choose Inc-v3 as the source model. As shown in Tab.~\ref{tab:combine_grad_opt_incv3}, though the combination of our method results in a marginal reduction in the white-box attack success rates, a notable increment is observed in the attack success rate within black-box conditions. Compared to other baseline methods, our method \name achieves higher attack success rates and surpasses the best baseline method by $4.7\%$, $3.5\%$, and $3.1\%$ for EMI, GMI, and VMI. These results highlight the good transferability and compatibility of our proposed method \name. 

\begin{table*}[htb]
    \caption{Attack success rates (\%) of various attacks when combined with three gradient optimization attacks, \ie, EMI, GMI, and VMI. The adversarial examples are generated on Inc-v3. * indicates the white-box model. The best results are in \textbf{bold}.}
    \centering
    \resizebox{0.9\linewidth}{!}{
    \begin{tabular}{c*{8}{b}}
        \toprule
         Attack & Inc-v3 & Inc-v4 & IncRes-v2 & Res-152 & VGG-16 & Inc-v3$_\mathrm{ens3}$ & Inc-v3$_\mathrm{ens4}$ & IncRes-v2$_\mathrm{ens}$\\
         \midrule
         EMI & \textbf{100.0*} & 70.1 & 66.0 & 52.2 & 56.5 & 21.9 & 23.7 & 11.1 \\
         FIA-EMI & ~~98.3* & 85.1 & 82.2 & 75.7 & 75.4 & 39.8 & 41.3 & 22.7 \\
         RPA-EMI & ~~97.9* & 84.2 & 84.0 & 79.3 & 76.4 & 47.9 & 49.5 & 28.7 \\
         NAA-EMI & ~~96.1* & 83.2 & 81.2 & 75.3 & 75.6 & 54.2 & 56.1 & 36.4 \\
         \name-EMI & ~~98.3* & \setrow{\bfseries}87.1 & 87.1 & 80.3 & 82.1 & 56.2 & 57.0 & 37.3\clearrow \\
         \midrule
         GMI & \textbf{100.0*} & 51.5 & 48.6 & 38.5 & 47.8 & 15.7 & 16.9 & 7.7 \\
         FIA-GMI & ~~98.9* & 87.5 & 84.3 & 79.0 & 76.9 & 34.4 & 36.3 & 18.1 \\
         RPA-GMI & ~~98.4* & 87.8 & 85.3 & 80.1 & 78.6 & 42.1 & 43.7 & 23.8 \\
         NAA-GMI & ~~97.7* & 85.9 & 83.8 & 77.8 & 75.9 & 48.4 & \setrow{\bfseries}50.1 & 31.7\clearrow \\
         \name-GMI & ~~97.9* & \setrow{\bfseries}88.3 & 87.8 & 81.7 & 81.1 & 50.2\clearrow & 49.8 & 31.2 \\
         \midrule
         VMI & \textbf{100.0*} & 68.8 & 66.5 & 55.6 & 57.3 & 31.5 & 34.5 & 19.6 \\
         FIA-VMI & ~~98.1* & 87.1 & 83.8 & 78.3 & 78.2 & 50.2 & 51.4 & 31.9 \\
         RPA-VMI & ~~97.8* & 87.4 & 85.2 & 80.6 & 79.8 & 58.6 & 58.4 & 39.8 \\
         NAA-VMI & ~~96.5* & 85.3 & 83.4 & 77.6 & 77.6 & 60.0 & 60.5 & 45.8 \\
         \name-VMI & ~~98.3* & \setrow{\bfseries}88.5 & 86.4 & 81.0 & 81.8 & 64.7 & 64.7 & 47.2 \clearrow\\
         \bottomrule
    \end{tabular}}
    \label{tab:combine_grad_opt_incv3}
\end{table*}

\textbf{Evaluation on input transformation-based attacks.}
We perform more comprehensive experiments on the baseline methods and our method \name when combined with three input transformation-based methods, including DIM, Admix, and SSA. The experimental setting is kept the same as in the main body. We test the attack success rates of the adversarial samples generated on Inc-v4, Res-152, and VGG-16 models. The results are respectively illustrated in Fig.~\ref{fig:combine_trans_incv4}, \ref{fig:combine_trans_res}, and \ref{fig:combine_trans_vgg}. We can observe that \name consistently achieves comparable performance with the baseline methods in most cases. Specifically, when utilizing the VGG-16 model as the source model and IncRes-v2$_\mathrm{ens}$ as the target model, our method achieves higher attack success rates, with an improvement margin ranging from 1.9\% to 2.2\% over the best-performing baseline method. Considering that the IncRes-v2$_\mathrm{ens}$ model has the best robustness among all the target models, this is a good indication of the effectiveness of our approach. Such superior results further validate \name's good compatibility with various input transformation-based attacks and deep models.

\section{SAFER vs SSA}
Although \name and SSA~\citep{long2022frequency} use similar strategies to perturb images in the frequency domain, there are still distinctions between them in three crucial aspects. \textbf{Motivation}. In contrast to SSA, which aims to yield diverse spectrum saliency maps, \name intends to disrupt the semantic and abstract features to calculate the weight matrix for highlighting crucial features. \textbf{Method}. Different from the approach of SSA, which introduces Gaussian noise and scales the spectrum randomly, our method \name employs \textsc{Self-Mix} within the frequency domain and applies \textsc{BlockMix} to the benign image during the computation of the weight matrix. This weight matrix serves as a fundamental component in constructing a novel feature-level loss for crafting more transferable adversarial examples. \textbf{Performance}. \name outperforms SSA with a margin of $11.1\%$ and can be combined with SSA to achieve better performance as in Fig.~\ref{fig:combine_trans}~(c).

\input{figs/transformation_incv4}
\begin{figure*}[tb]
    \centering
    \begin{subfigure}{\linewidth}
    \centering
        \begin{subfigure}{0.18\linewidth}
            \includegraphics[width=0.9\linewidth]{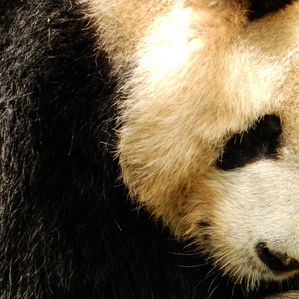}
            \caption*{Original Image}
        \end{subfigure}
        \begin{subfigure}{0.18\linewidth}
            \includegraphics[width=0.9\linewidth]{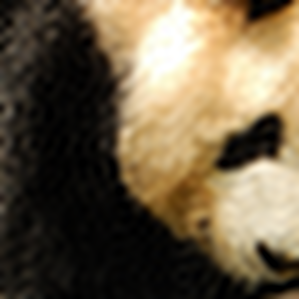}
            \caption*{L-Freq. 0-40}
        \end{subfigure}
        \begin{subfigure}{0.18\linewidth}
            \includegraphics[width=0.9\linewidth]{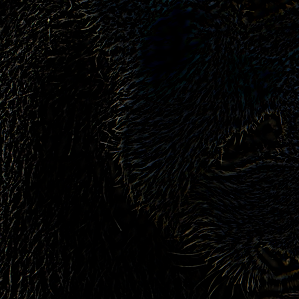}
            \caption*{H-Freq. 40-299}
        \end{subfigure}
        \begin{subfigure}{0.18\linewidth}
            \includegraphics[width=0.9\linewidth]{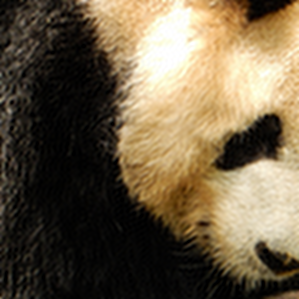}
            \caption*{L-Freq. 0-100}
        \end{subfigure}
        \begin{subfigure}{0.18\linewidth}
            \includegraphics[width=0.9\linewidth]{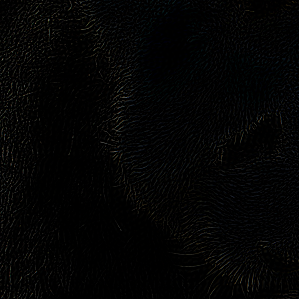}
            \caption*{H-Freq. 100-299}
        \end{subfigure}
        \\
        \begin{subfigure}{0.18\linewidth}
            \includegraphics[width=0.9\linewidth]{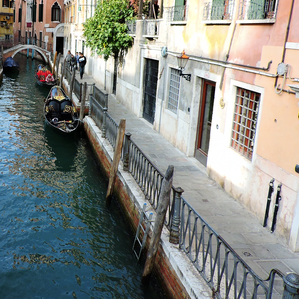}
            \caption*{Original Image}
        \end{subfigure}
        \begin{subfigure}{0.18\linewidth}
            \includegraphics[width=0.9\linewidth]{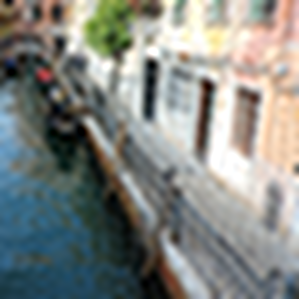}
            \caption*{L-Freq. 0-40}
        \end{subfigure}
        \begin{subfigure}{0.18\linewidth}
            \includegraphics[width=0.9\linewidth]{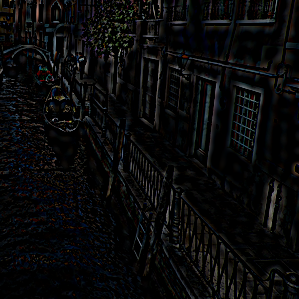}
            \caption*{H-Freq. 40-299}
        \end{subfigure}
        \begin{subfigure}{0.18\linewidth}
            \includegraphics[width=0.9\linewidth]{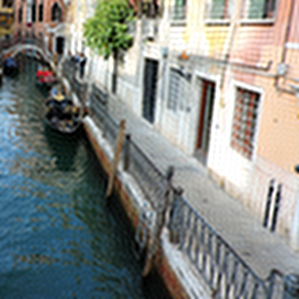}
            \caption*{L-Freq. 0-100}
        \end{subfigure}
        \begin{subfigure}{0.18\linewidth}
            \includegraphics[width=0.9\linewidth]{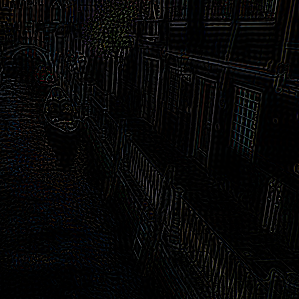}
            \caption*{H-Freq. 100-299}
        \end{subfigure}
        \\
        \begin{subfigure}{0.18\linewidth}
            \includegraphics[width=0.9\linewidth]{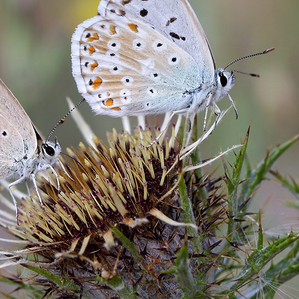}
            \caption*{Original Image}
        \end{subfigure}
        \begin{subfigure}{0.18\linewidth}
            \includegraphics[width=0.9\linewidth]{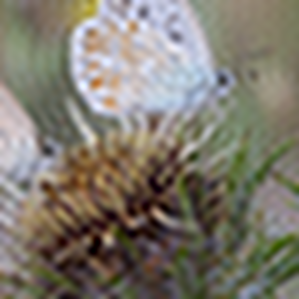}
            \caption*{L-Freq. 0-40}
        \end{subfigure}
        \begin{subfigure}{0.18\linewidth}
            \includegraphics[width=0.9\linewidth]{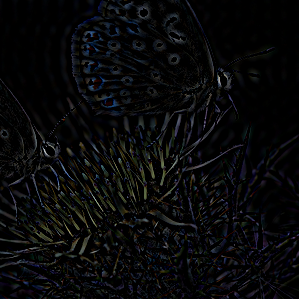}
            \caption*{H-Freq. 40-299}
        \end{subfigure}
        \begin{subfigure}{0.18\linewidth}
            \includegraphics[width=0.9\linewidth]{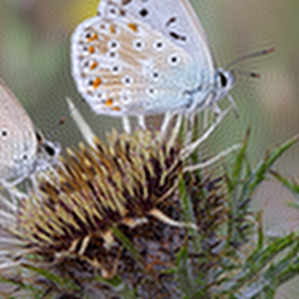}
            \caption*{L-Freq. 0-100}
        \end{subfigure}
        \begin{subfigure}{0.18\linewidth}
            \includegraphics[width=0.9\linewidth]{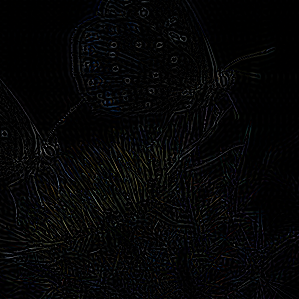}
            \caption*{H-Freq. 100-299}
        \end{subfigure}
        \\
        \begin{subfigure}{0.18\linewidth}
            \includegraphics[width=0.9\linewidth]{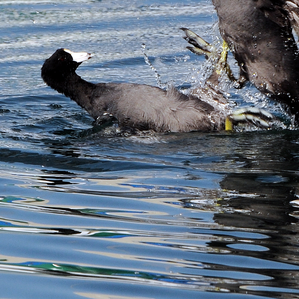}
            \caption*{Original Image}
        \end{subfigure}
        \begin{subfigure}{0.18\linewidth}
            \includegraphics[width=0.9\linewidth]{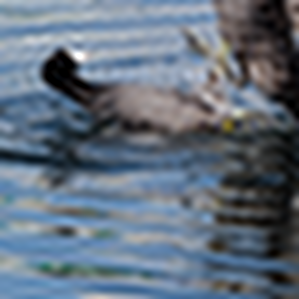}
            \caption*{L-Freq. 0-40}
        \end{subfigure}
        \begin{subfigure}{0.18\linewidth}
            \includegraphics[width=0.9\linewidth]{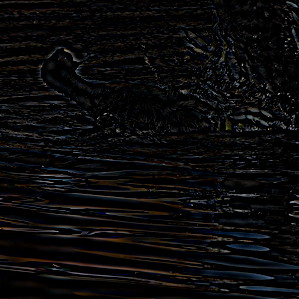}
            \caption*{H-Freq. 40-299}
        \end{subfigure}
        \begin{subfigure}{0.18\linewidth}
            \includegraphics[width=0.9\linewidth]{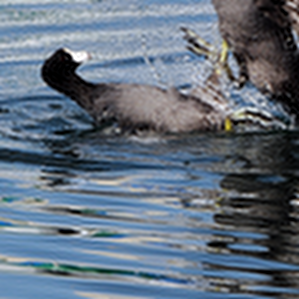}
            \caption*{L-Freq. 0-100}
        \end{subfigure}
        \begin{subfigure}{0.18\linewidth}
            \includegraphics[width=0.9\linewidth]{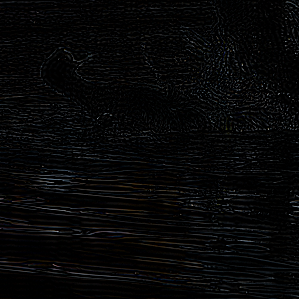}
            \caption*{H-Freq. 100-299}
        \end{subfigure}
        \\
        \begin{subfigure}{0.18\linewidth}
            \includegraphics[width=0.9\linewidth]{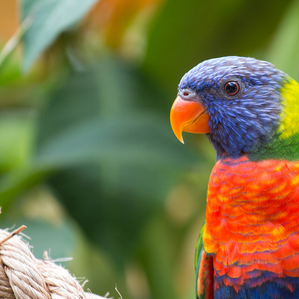}
            \caption*{Original Image}
        \end{subfigure}
        \begin{subfigure}{0.18\linewidth}
            \includegraphics[width=0.9\linewidth]{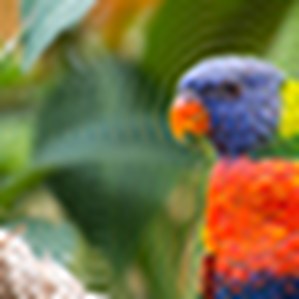}
            \caption*{L-Freq. 0-40}
        \end{subfigure}
        \begin{subfigure}{0.18\linewidth}
            \includegraphics[width=0.9\linewidth]{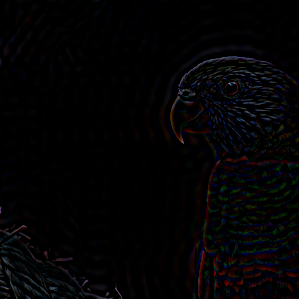}
            \caption*{H-Freq. 40-299}
        \end{subfigure}
        \begin{subfigure}{0.18\linewidth}
            \includegraphics[width=0.9\linewidth]{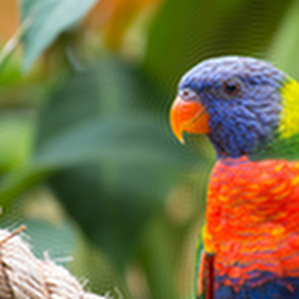}
            \caption*{L-Freq. 0-100}
        \end{subfigure}
        \begin{subfigure}{0.18\linewidth}
            \includegraphics[width=0.9\linewidth]{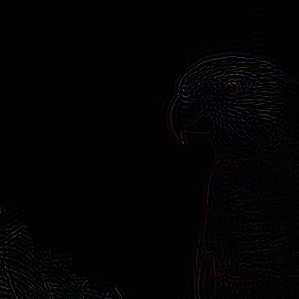}
            \caption*{H-Freq. 100-299}
        \end{subfigure}
    \end{subfigure}
    \caption{Visualization of high-frequency and low-frequency components, denoted as H-Freq and L-Freq, respectively.}
    \label{fig:freq_recover_visual}
\end{figure*}

\section{Visualisation}
In this study, we aim to balance perturbation between abstract and semantic features associated with high- and low-frequency components, respectively. To better reveal this relationship and provide insight into abstract and semantic content, we visualize these aspects by selectively showing either high- or low-frequency components. Given the input image dimension $299 \times 299 \times 3$, we designate the upper left square region within the frequency domains, spanning from $0$ to $\tau$, as low-frequency components, and the remainder as high-frequency components, indicated as $\tau$ to $299$, where we assign $\tau = 40, 100$. The results are shown in Fig.~\ref{fig:freq_recover_visual}. Notably, images recovered from the low-frequency band exhibit richer semantic information, whereas those recovered from the high-frequency band portray intricate details, \ie texture, edges \etc, posing perceptual challenges as frequency rises. Furthermore, it is observed that only a small portion of low-frequency components can yield relatively high-quality images, further confirming that low-frequency components occupy only a minor fraction of the entire frequency spectrum.

\end{document}